\documentclass[10pt,twocolumn,letterpaper]{article}

\usepackage{cvpr}              %

\usepackage{amsmath}            
\usepackage{amssymb}            
\usepackage{multirow}           
\usepackage{booktabs}              
\usepackage[linesnumbered,ruled,vlined]{algorithm2e}
\usepackage{amsmath}
\usepackage{amssymb}
\SetKwInput{KwInitial}{Initialize}
\usepackage{float}
\usepackage{subcaption} %
\usepackage{colortbl}

\newcommand{\ft}{\texttt{IM}}
\newcommand{\sss}{\texttt{S\textsuperscript{3}}}
\newcommand{\ims}{\texttt{ImS\textsuperscript{3}}}
\newcommand{\sectionnotoc}[1]{
  \refstepcounter{section}%
  \section*{\thesection\quad #1}%
}
\newcommand{\subsectionnotoc}[1]{%
  \refstepcounter{subsection}%
  \subsection*{\thesubsection\quad #1}%
}

\definecolor{cvprblue}{rgb}{0.21,0.49,0.74}
\usepackage[pagebackref,breaklinks,colorlinks,allcolors=cvprblue]{hyperref}
\usepackage{multirow}
\usepackage[accsupp]{axessibility}
\def\paperID{} %
\def\confName{CVPR}
\def\confYear{2026}

\title{IMS3: Breaking Distributional Aggregation in \\
Diffusion-Based Dataset Distillation}

\author{
Chenru Wang$^{1}$\thanks{These authors contributed equally.} \qquad
Yunyi Chen$^{1,2}$\footnotemark[1] \qquad
Zijun Yang$^{3}$ \qquad
Joey Tianyi Zhou$^{4,5}$ \qquad
Chi Zhang$^{1}$\thanks{Corresponding author.}\\
$^{1}$AGI Lab, Westlake University,~
$^{2}$Eindhoven University of Technology,~\\
$^{3}$Jinan University,~
$^{4}$IHPC, Agency for Science, Technology
and Research, Singapore \\
$^{5}$CFAR, Agency for Science, Technology and Research, Singapore \\
}

\makeatletter
\def\@fnsymbol#1{\ensuremath{\ifcase#1\or *\or \dagger\fi}}
\makeatother

\begin{document}
\maketitle

\begin{abstract}

Dataset Distillation aims to synthesize compact datasets that can approximate the training efficacy of large-scale real datasets, offering an efficient solution to the increasing computational demands of modern deep learning. Recently, diffusion-based dataset distillation methods have shown great promise by leveraging the strong generative capacity of diffusion models to produce diverse and structurally consistent samples. However, a fundamental goal misalignment persists: diffusion models are optimized for generative likelihood rather than discriminative utility, resulting in over-concentration in high-density regions and inadequate coverage of boundary samples crucial for classification. To address this issue, we propose two complementary strategies. Inversion-Matching (\ft) introduces an inversion-guided fine-tuning process that aligns denoising trajectories with their inversion counterparts, broadening distributional coverage and enhancing diversity. Selective Subgroup Sampling (\sss) is a training-free sampling mechanism that improves inter-class separability by selecting synthetic subsets that are both representative and distinctive. Extensive experiments demonstrate that our approach significantly enhances the discriminative quality and generalization of distilled datasets, achieving state-of-the-art performance among diffusion-based methods.

\end{abstract}
    
\sectionnotoc{Introduction}
\label{sec:intro}

\begin{figure}[t]
    \centering
    \includegraphics[width=1\linewidth]{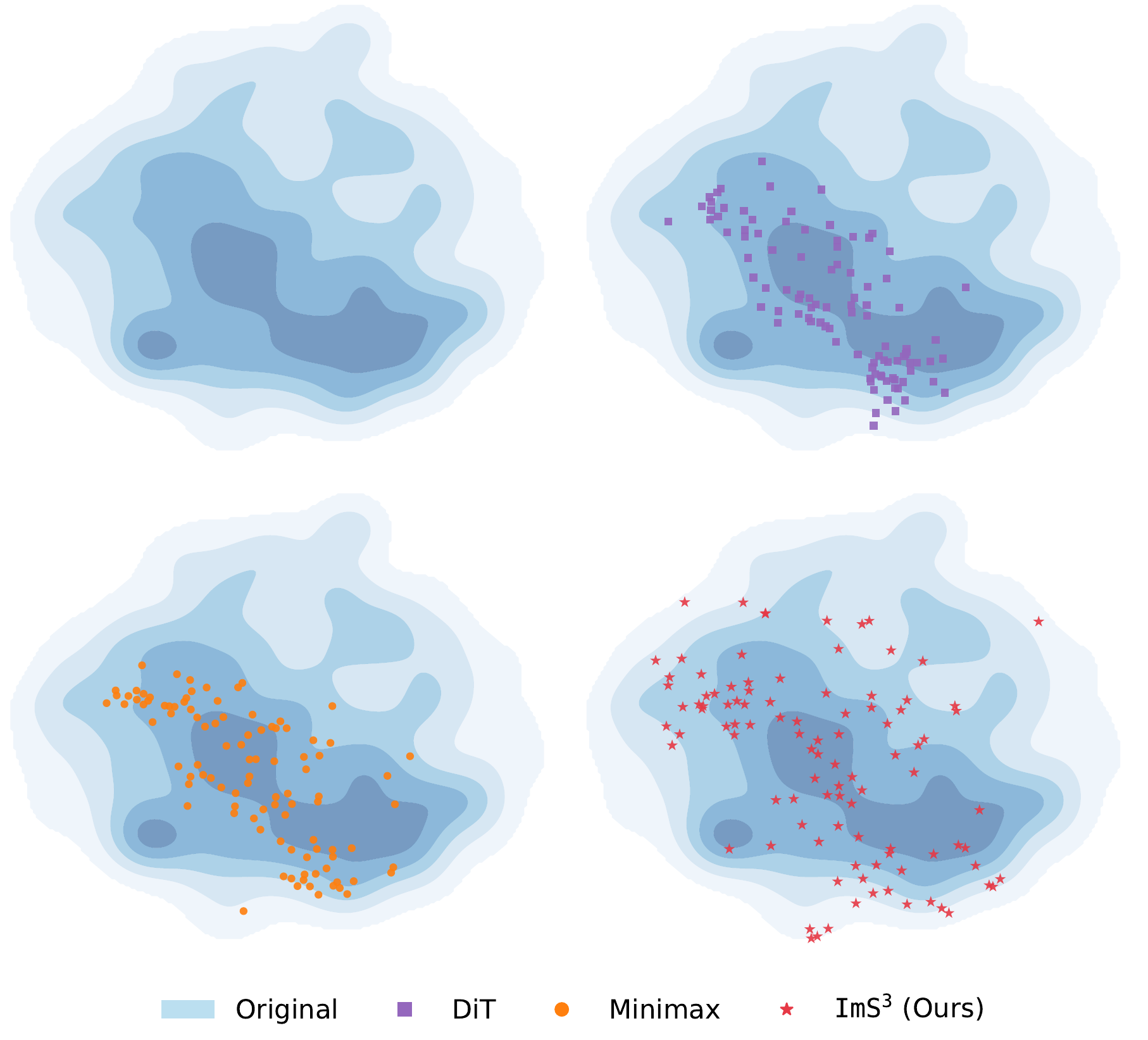}
    \caption{t-SNE visualization of feature distributions for different distilled datasets.
    The blue density map represents the feature distribution of the original dataset.
    The purple squares correspond to samples generated by the DiT~\cite{DiT}, the orange dots correspond to Minimax~\cite{minimax}, and the red stars indicate samples produced by our~\ims.
    Our method covers a broader area of the feature space, indicating improved distribu-
    tional coverage. }
    \label{fig:tsne}
\end{figure}

The rapid growth of deep learning has been accompanied by an equally rapid increase in dataset size and computational demands~\cite{DL,imagenet15russakovsky}. To reduce training costs and memory requirements, researchers have explored dataset reduction strategies. Early approaches often relied on data pruning, which selects a subset of the original data but suffers from a limited compression ratio. Recently, Dataset Distillation has emerged as a more effective paradigm~\cite{wang2018dataset,MTT,GM,survey1,survey2,dd1,dd2,r31,r32}. Instead of directly selecting from the original dataset, distillation generates a compact synthetic dataset that can approximate the training performance of the full dataset. This generative perspective enables substantially higher compression ratios while maintaining competitive accuracy, making dataset distillation an attractive solution in modern large-scale learning scenarios.

Earlier works mainly formulated dataset distillation as an optimization problem, where a small set of synthetic samples is directly optimized to match the statistical or gradient information of the original dataset~\cite{MTT,GM,IDC,opt1,opt2,opt3}. However, such optimization-based methods often suffer from high computational costs and poor scalability to high-resolution or large-class datasets due to their reliance on iterative gradient alignment and sample-specific updates.
Recently, with the emergence and widespread adoption of diffusion models~\cite{ddpm,DDIM,SMSED,wang2026freelunchstabilizingrectified}, diffusion-based dataset distillation methods have emerged and been widely adopted~\cite{minimax,D4M,cao2,D3HR,ddvlcp,mgd3,dbdd1,dbdd2,dbdd3}. By leveraging the generative capacity of diffusion models,%
diffusion-based methods benefit from their probabilistic formulation, which allows them to capture complex data manifolds and generate samples with higher stability and structural consistency~\cite{ELBO}, %
these methods can synthesize high-quality and diverse samples while preserving the underlying structure of the original dataset. %
Typically, diffusion-based distillation consists of two stages, fine-tuning and sampling, which respectively adapt the generative model and determine how distilled samples are drawn. %
Compared with optimization-based methods, diffusion-based distillation offers improved fidelity, scalability, and robustness, making it a promising direction for large-scale applications.

Despite these advances, existing diffusion-based dataset distillation methods face a fundamental goal misalignment between generation and discrimination. A diffusion model is inherently generative, trained to maximize data likelihood and thereby concentrate on high-density regions of the data manifold~\cite{biroli2024dynamical,ddpm,karczewski2025diffusionmodelscartoonistscurious}. In contrast, dataset distillation serves a discriminative objective, where effective dataset distillation requires capturing critical samples near decision boundaries that are often low-density and hard to learn.
This mismatch leads to what we refer to as distributional aggregation, where synthetic data are overly concentrated in high-density areas while underrepresenting boundary regions crucial for generalization and robust discrimination.
As illustrated in \cref{fig:tsne}, diffusion-based methods such as Minimax~\cite{minimax} exhibit a strong aggregation of samples in high-density regions of the feature space, whereas our proposed~\ims produces more uniformly distributed features, indicating improved coverage and representation diversity.

To address this issue, we aim to mitigate the model’s inherent high-density bias and improve the representational diversity of synthetic datasets.
To achieve this goal, we exploit a distinctive property of the diffusion inversion process: its inherent instability causes inversion trajectories to deviate from high-density regions and naturally shift toward low-density areas of the data manifold~\cite{instability,NTI}. Building upon this insight, we propose Inversion-Matching (\ft), a novel fine-tuning strategy, which 
explicitly aligns the noise latent with its inversion counterpart at corresponding timestep.
This alignment serves as a corrective signal that encourages the model to broaden its distributional coverage, fill underrepresented regions, and enhance sample diversity, thereby producing more discriminative and representative distilled datasets.

Standard diffusion sampling procedures are primarily designed for generative quality rather than discriminative utility~\cite{DDIM,dpm,dpmpp}. They typically generate data for each class independently, without explicitly considering inter-class relationships. As a result, the synthesized samples, though visually plausible, may lead to reduced discriminative power in downstream classifiers.
To explicitly enhance the class separability of distilled datasets, we propose Selective Subgroup Sampling (\sss), a training-free inference-time sampling strategy tailored to improve discriminative representation. Specifically, during the generation of each class,~\sss~produces multiple candidate sample groups in the feature space and evaluates them based on two complementary criteria: proximity to the real data centroid (ensuring representativeness) and separation from other class centroids (ensuring distinction). The candidate group that best satisfies both criteria is selected as the distilled subset. This simple yet effective mechanism introduces class awareness into the sampling process without any additional training cost. By encouraging each class’s synthetic samples to be both representative and well-separated,~\sss~enhances the inter-class discrimination of the distilled dataset and significantly improves the downstream classification performance.

In summary, we revisit the key challenges of dataset distillation and address them from two complementary perspectives.~\ft~expands the coverage of synthetic data by guiding the diffusion model toward low-density regions, thereby mitigating distributional aggregation and improving representation diversity.~\sss~further enhances the discriminative structure of the distilled dataset by explicitly promoting inter-class separation during sampling.
To validate the effectiveness of our strategy, we conduct extensive experiments across multiple datasets and model architectures. The results demonstrate that our method consistently achieves superior accuracy compared with existing diffusion-based distillation approaches, confirming the effectiveness of the proposed fine-tuning and sampling strategies.

Our contributions are summarized as follows:
\begin{itemize}
\item We propose~\ft~to mitigate distributional aggregation and improve coverage of low-density regions.
\item We introduce~\sss, a training-free sampling strategy that enhances inter-class discriminability.
\item Extensive experiments demonstrate that our approach achieves state-of-the-art performance on multiple datasets. Code and model will be publicly released to facilitate future research.
\end{itemize}

 \sectionnotoc{Related Works}
\label{sec:related}

\subsectionnotoc{Non-generative Dataset Distillation}
Early studies on dataset distillation mainly adopt optimization-based paradigms, where a compact synthetic set is directly optimized to approximate the training dynamics of real data.
The seminal work Dataset Distillation~\cite{wang2018dataset} formulates this task as a bi-level optimization problem that minimizes the gradient mismatch between real and synthetic batches.
To improve efficiency and scalability, $\mathrm{SRe^2L}$~\cite{Sre2L} and its following works~\cite{se1,se2,zhou2024self} adopt a uni-level optimization framework that learns shared representations and relaxed labels, significantly reducing computation while maintaining performance.
Other representative approaches, such as Matching Training Trajectories (MTT)~\cite{MTT} and Gradient Matching~\cite{GM}, further improve gradient alignment or optimize learning trajectories.
Although these methods achieve promising results on small-scale datasets, they generally struggle to scale to high-resolution or large-class scenarios due to the inherent optimization burden and limited generative diversity.

\subsectionnotoc{Diffusion-based Dataset Distillation}

Recent studies have leveraged diffusion models for dataset distillation to synthesize compact yet representative datasets.
A representative work, Minimax Diffusion~\cite{minimax}, fully utilizes pre-trained diffusion models for generating realistic synthetic datasets and fine-tunes them with a minimax criterion to enhance both sample representativeness and diversity.
Following this line, $\mathrm{D^4M}$~\cite{D4M} adopts pre-trained text-to-image diffusion models and integrates prototype learning to cluster latent features and learn category centers, thereby improving intra-class consistency and visual fidelity.
To address distribution mismatch and randomness in latent sampling, $\mathrm{D^3HR}$~\cite{D3HR} introduces a domain-mapping framework that employs DDIM inversion to map VAE latents into a Gaussian domain with higher normality, coupled with a group-sampling strategy for improved representativeness and distribution alignment.
More recently, DDVLCP~\cite{ddvlcp} extends diffusion-based distillation into the vision–language domain by constructing paired image-text prototypes via large language models, effectively enhancing semantic coherence and mitigating illogical or missing-object artifacts during generation.
While these methods markedly improve fidelity and scalability over optimization-based approaches, they still exhibit distributional aggregation, where generated samples tend to cluster in high-density regions, reducing diversity and coverage of fine-grained low-density patterns.

\subsectionnotoc{Inversion in Diffusion Models}

In diffusion models, inversion aims to trace a real image back to its originating noise. While the reverse process can directly add noise and reach the initial noise state in a single reverse step, inversion must recover the latent trajectory through step-by-step estimation of the intermediate states.
This technique has been widely used in diffusion-based editing, alignment, watermark, and reconstruction~\cite{DDIM,NTI,pnp,rf-inversion,rf-solver,fireflow}, providing a means to bridge real and generative data domains.

However, recent studies have shown that inversion process has inherent instability due to the compounding approximation error along the inversion trajectory~\cite{instability}.
Since the forward and inverse dynamics are not perfectly symmetric, small local errors can accumulate and push the inversion path away from high-density manifolds of the real data distribution. 
Consequently, inverted latents often drift toward low-density regions of the data space, forming an implicit bias that reflects structural inconsistencies between the true and modeled distributions.
While this instability is generally undesirable in reconstruction tasks, our work leverages it as a beneficial property to intentionally expand the coverage of low-density areas during fine-tuning, thereby mitigating the over-aggregation problem in diffusion models.

\begin{figure*}[t]
    \centering
    \includegraphics[width=\linewidth]{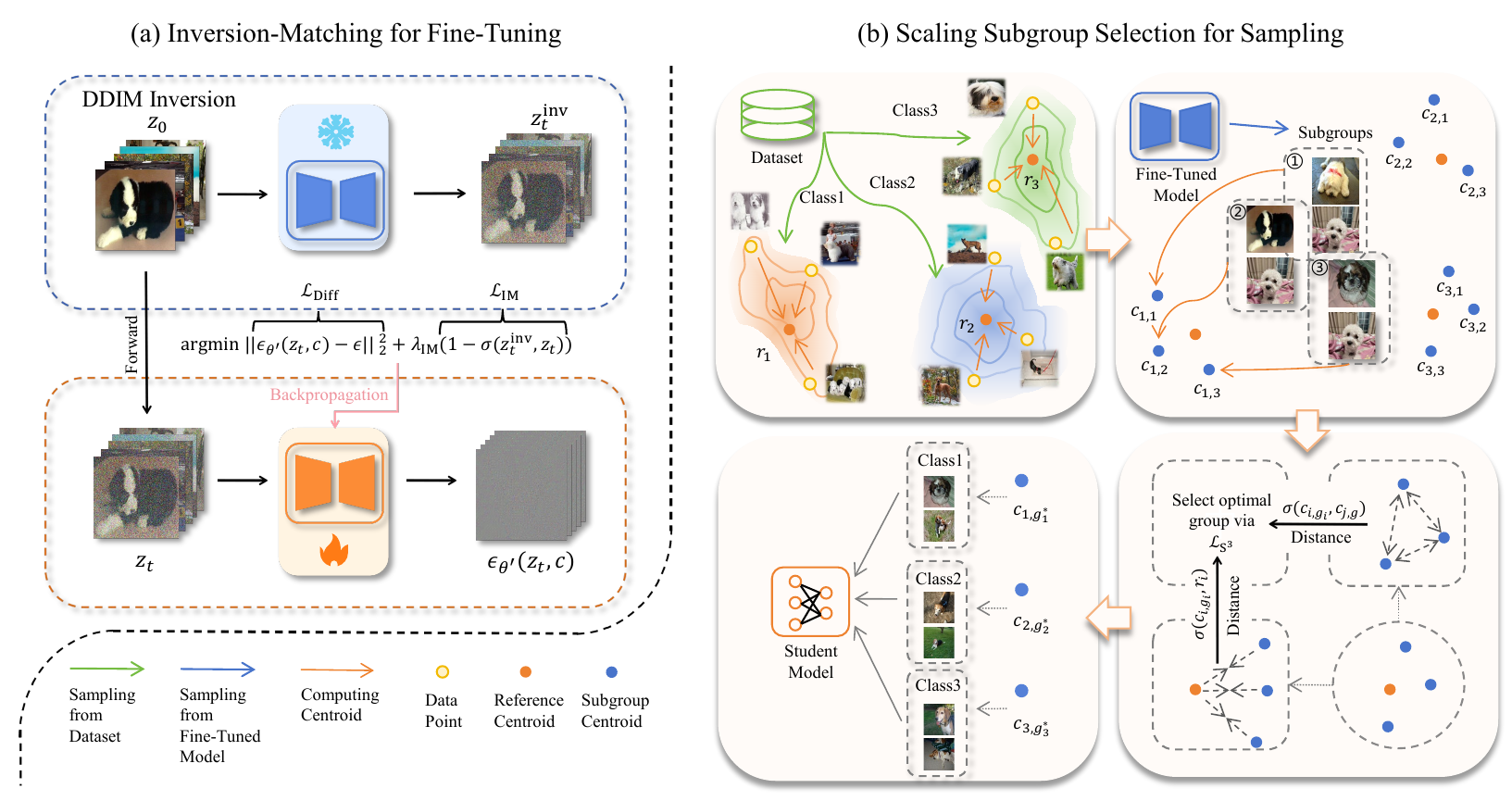}
    \caption{Overview of the proposed~\ims~framework. 
(a)~\ft~fine-tuning expands distribution coverage by aligning denoised and inverted latents through the~\ft~loss~$\mathcal{L}_{\mathrm{IM}}$. 
(b)~\sss~enhances discriminative diversity during sampling by selecting representative subgroups whose centroids are close to real data centroids while maintaining inter-class separation, optimized via~$\mathcal{L}_{\mathrm{S}^3}$. 
$r_i$ denotes the centroid of real samples from the $i$-th class, while $c_{i,g_i}$ represents the centroid of the $g_i$-th candidate subgroup synthesized by the diffusion model for the same class.
The selected optimal subgroup centroid $c_{i,g_i^*}$ strikes the best balance between representativeness to the real class distribution and distinctiveness from other classes.
}
    \label{fig:pipeline}
\end{figure*}

\sectionnotoc{Preliminary}
\label{sec:preliminary}

Before introducing our methodology, we first provide some necessary background information in this section.

\subsectionnotoc{Dataset Distillation}
Dataset distillation aims to construct a compact synthetic dataset that can approximate the training efficacy of the original large-scale dataset.
Given an original dataset $\mathcal{D}_{r}=\{(x_i,y_i)\}_{i=1}^{N}$ with $C_{cls}$ classes, the goal is to synthesize a much smaller distilled dataset $\mathcal{D}_{s}=\{(\tilde{x}_j,\tilde{y}_j)\}_{j=1}^M$, where $M \ll N$.
A model trained on $\mathcal{D}_{s}$ should achieve comparable or even higher performance to one trained on $\mathcal{D}_{r}$ when evaluated on unseen real data.

In practice, the effectiveness of dataset distillation is typically measured under the \emph{Images Per Class} (IPC) setting, where each class in $\mathcal{D}_{s}$ contains a limited number of samples (\emph{e.g.}, 1, 10, or 50).
The central challenge lies in generating or selecting these few samples such that they preserve both the coverage and the discriminative structure of the original distribution, enabling models trained on the distilled data to generalize effectively.

\subsectionnotoc{DDIM Inversion}
Consider a deterministic DDIM sampler with $\eta=0$ and cumulative noise schedule $\bar\alpha_t\in(0,1]$. The forward denoising update can be written as
\begin{equation}
\label{eq:ddim-forward}
\mathbf{z}_{t-1} \;=\; \sqrt{\bar\alpha_{t-1}}\hat{\mathbf{z}}_0(z_t,t)\;+\;\sqrt{1-\bar\alpha_{t-1}}\;\varepsilon_\theta(\mathbf{z}_t,t),
\end{equation}
where
\begin{equation}
\hat{\mathbf{z}}_0(\mathbf{z}_t,t)\;=\;\frac{\mathbf{z}_t-\sqrt{1-\bar\alpha_t}\,\varepsilon_\theta(\mathbf{z}_t,t)}{\sqrt{\bar\alpha_t}}.
\end{equation}
DDIM inversion (Euler scheduler) performs the reverse of the sampling process, mapping a real image into its latent noise. Based on the DDIM update rule, the inversion step of Euler scheduler using approximation is given by:  
\begin{equation}
\label{eq:lim}
    \mathbf{z}_{t_{i}} =\mathbf{z}_{t_{i-1}}+(\sigma_{{t_{i}}}-\sigma_{{t_{i-1}}})\varepsilon_{\theta}(\mathbf{z}_{{t_{i-1}}},{t_{i-1}}),
\end{equation}
where $\sigma_{{t_{i}}} $ and $\sigma_{{t_{i-1}}}$ are scheduling parameters.

Theoretically, reconstructions along inversion trajectories tend to occupy low-density regions of the data manifold due to the inherent instability of the inversion process~\cite{instability}, as discussed in \cref{sec:related}.
\emph{In this work, we exploit this property during fine-tuning to intentionally guide the generative model toward underrepresented, low-density areas, which will be further elaborated in \cref{sec:im}.} A more detailed introduction of the inversion instability is provided in the Appendix.

\sectionnotoc{Method}
\label{sec:method}

In this section, we introduce our diffusion-based dataset distillation strategies, termed \ims.
The proposed method aims to construct a compact yet representative synthetic dataset by enhancing both distributional coverage and discriminative diversity.
The overall pipeline of \ims~is illustrated in \cref{fig:pipeline}.
It consists of two stages: Inversion-Matching (\ft) fine-tunes a pretrained diffusion model to expand low-density coverage by aligning denoised and inverted latents, while Selective Subgroup Sampling (\sss) enhances discriminative diversity during sampling by selecting representative subgroups based on centroid similarity.

\subsectionnotoc{Inversion-Matching for Fine-Tuning}
\label{sec:im}

Diffusion-based dataset distillation typically involves fine-tuning a pre-trained generative model on the target dataset, followed by using the adapted model to synthesize a compact distilled dataset. However, this process often suffers from distributional aggregation, where generated samples concentrate in high-density regions of the data manifold, leaving low-density regions underrepresented, where decision boundaries usually lie and which are essential for effective dataset distillation.

To mitigate this imbalance, we develop~\ft, a fine-tuning strategy that leverages the intrinsic property of inversion, whose inherent instability causes the inversion trajectories to deviate from high-density regions, guiding the generative model toward broader coverage of the data space. Here, we choose the DiT-XL/2~\cite{DiT} as the pre-trained diffusion model. This method can be decomposed into the following two parts.
 
\noindent\textbf{Training Objective.}
To enable the model to learn a distribution with broader coverage, we first follow the deterministic DDIM inversion in \cref{eq:lim} to compute the inversion noise latent $\mathbf{z}^{\mathrm{inv}}_t$ at timestep $t$.
For efficiency, instead of reconstructing the denoised latent by running the entire denoising trajectory from $T$ to $t$, we directly sample the noise latent at timestep $t$ by
\begin{equation}
    \mathbf{z}_t = \sqrt{\bar{\alpha}_t}\,\mathbf{z}_0 + \sqrt{1-\bar{\alpha}_t}\,\boldsymbol{\epsilon},
\end{equation}
where $\boldsymbol{\epsilon}\!\sim\!\mathcal{N}(0,\mathbf{I})$.
We then perform time-aligned~\ft, that is, matching the noise latent and inverted noise latent at the same timestep $t$ using
\begin{equation}
\label{eq:im-loss-timealigned}
\mathcal{L}_{\mathrm{IM}}
= 1 - \sigma(\,\mathbf{z}^{\mathrm{inv}}_t,\mathbf{z}_t),
\end{equation}
where $\sigma(\cdot,\cdot)$ denotes cosine similarity.
This objective aligns the noise latent $\mathbf{z}_t$ with the inverted noise latent $\mathbf{z}^{\mathrm{inv}}_t$ without enforcing exact equality between the two latent states, thereby guiding the model to capture underrepresented regions of the data manifold.

Additionally, in order to prevent the excessive distributional shift that can degrade the fidelity of generated images, we include the standard diffusion loss to minimize the squared error between the predicted noise $\epsilon_\theta(\mathbf{z}_t,\,c)$ and the ground truth $\epsilon$:
 
\begin{equation}
\label{eq:ldiff}
\mathcal{L}_{\mathrm{Diff}}
=\left\|
\epsilon_\theta(\mathbf{z}_t,\,c) - \epsilon
\right\|_2^2.
\end{equation}

The overall fine-tuning loss combines high-density fidelity and low-density coverage:
\begin{equation}\label{loss}
\mathcal{L}
=
\mathcal{L}_{\mathrm{Diff}}
\;+\;
\lambda_{\mathrm{IM}}\,\mathcal{L}_{\mathrm{IM}},
\end{equation}
where $\lambda_{\mathrm{IM}}>0$ balances the two terms~(analysis see Appendix).

Full fine-tuning of $\mathcal{G}_\theta$ is computationally expensive and memory-intensive. Following Minimax Diffusion~\cite{minimax} and for efficiency, we adopt a Parameter Efficient Fine-Tuning (PEFT) approach via Difffit~\cite{difffit}, which inserts lightweight adapters into attention/MLP blocks and updates only these small modules while freezing the backbone, thereby reducing trainable parameters and compute without sacrificing stability or fidelity. 

\cref{alg:im} outlines the fine-tuning stage, where noise latent and inverted noise latent are aligned through a cosine-similarity loss. Combined with the standard diffusion loss. Overall, our fine-tuning strategy enables efficient adaptation of the diffusion model without introducing significant computational overhead, while effectively expanding distributional coverage and preserving generation fidelity.

\begin{algorithm}[tb]
\caption{Inversion-Matching~(\ft)}
\label{alg:im}
\SetAlgoLined
\SetKwInput{KwIn}{Input}
\SetKwInput{KwOut}{Output}
\SetKwInput{KwInitial}{Initialize}

\KwIn{Pre-trained parameters $\theta$, encoder $E$, class encoder $E_c$, real dataset $\mathcal{D}_r$, feature extractor $\phi(\cdot)$}
\KwOut{Fine-tuned generator $\mathcal{G}_{\theta'}$}
\KwInitial{${\theta}^{'} \leftarrow \theta$}
\BlankLine
\For{each real sample $(x_0,y_0) \sim \mathcal{D}_r$}{
    Sample timestep $t$ and noise $\boldsymbol{\epsilon}\sim\mathcal{N}(0,\mathbf{I})$\;
    Compute latent $z_0 \leftarrow E(x_0)$ and class embedding $c \leftarrow E_c(y_0)$\;
    Obtain inversion state $z_t^{\mathrm{inv}} \leftarrow \text{Inversion}(z_0, t, \theta^{'})$ 
    Compute noisy latent: $\displaystyle z_t \leftarrow \sqrt{\bar{\alpha}_t}\,z_0 + \sqrt{1-\bar{\alpha}_t}\,\boldsymbol{\epsilon}$\;
    Compute diffusion loss: $\displaystyle \mathcal{L}_{\mathrm{Diff}}=\big\|\epsilon_{{\theta}^{'}}(z_t,c)-\boldsymbol{\epsilon}\big\|_2^2$\;
    Compute inversion-matching loss: $\displaystyle \mathcal{L}_{\mathrm{IM}} = 1-\sigma\!\left(z_t^{\mathrm{inv}},\,z_t\right)$\;
    Update the model parameter ${\theta}^{'}$ with Eq.~\eqref{loss}\;
}
Set $\mathcal{G}_{\theta'}$ as the fine-tuned generator\;

\end{algorithm}

\subsectionnotoc{Selective Subgroup Sampling}
\label{sec:sss}

Standard diffusion sampling primarily focuses on generative fidelity, often ignoring inter-class relationships among categories.
As a result, the synthesized data may lack sufficient discriminative structure, which is critical for effective dataset distillation.

To address this issue, we propose~\sss, a training-free sampling strategy that aims to ensure the distilled samples in each class are close to the real samples of the same class while remaining far from the distilled samples of other classes. 
To achieve this,~\sss~selects representative and diverse subgroups based on centroid similarity. 

\noindent\textbf{Reference Centroids.}
\label{subsec: ref-cen}
To measure and align the differences between real and synthetic distributions, thereby achieving our goal of balancing coverage and discrimination, we employ a cluster centroid representation to quantify distributional relations.

Formally, let $\mathcal{D}_{r}$ be a labeled dataset with $C$ classes in the distilled setting. A frozen feature encoder $\phi:\mathbb{R}^{H\times W\times 3}\!\to\!\mathbb{S}^{C_\mathrm{Channel}}$ maps images to $\ell_2$-normalized embeddings on the unit sphere. For class $i$, denote the real subset with $K_i$ samples by $\mathcal{D}_i$~(analysis see \cref{subsec:k}), and define the real class centroid
\begin{equation}
\label{eq:real-centroid}
r_i \;=\; \big \|\frac{1}{K_i}\sum_{x\in\mathcal{D}_i}\phi(x)\big\|_2\in\mathbb{S}^{C_\mathrm{Channel}}.
\end{equation}

\noindent\textbf{Candidate Subgroups and Centroids.}
Following the definition aforementioned, we compute the centroids of the subgroups.
Suppose that $\mathcal{C}=\{1,2,\cdots,C\}$ and $\mathcal{G}=\{1,2,\cdots,G\}$. 
For each class $i \in \mathcal{C}$, we draw $G$ candidate subgroups
$\{\mathcal{S}_{i,g_i}\}_{g_i \in \mathcal{G}}$ from a pretrained diffusion generator $\mathcal{G}_\theta$~(\emph{e.g.}, DiT~\cite{DiT})~and a frozen VAE decoder~(analysis of $G$ see Appendix);
each subgroup contains $K$ images.
All candidate images are embedded by $\phi$, and each subgroup centroid is defined as
\begin{equation}
\label{eq:subgroup-centroid}
c_{i,g_i}\;=\;\Big\|\frac{1}{K}\!\sum_{x\in\mathcal{S}_{i,g_i}}\phi(x)\Big\|_2
\in\mathbb{S}^{C_\mathrm{Channel}}.
\end{equation}

After obtaining all subgroup centroids, we evaluate them based on their representativeness and discriminative separation to identify the optimal subset for each class.

\begin{algorithm}[tb]
\caption{Selective Subgroup Sampling (\sss)}
\label{alg:s3}
\SetAlgoLined
\SetKwInput{KwIn}{Input}
\SetKwInput{KwOut}{Output}
\SetKwInput{KwInitial}{Initialize}

\KwIn{Fine-tuned generator $\mathcal{G}_{\theta'}$, real dataset $\mathcal{D}_r=\{\mathcal{D}_i\}_{i=1}^{C}$, feature extractor $\phi(\cdot)$, number of candidates $G$}
\KwOut{Distilled dataset $\mathcal{D}_s=\{S_i^*\}_{i=1}^{C}$}
\KwInitial{Randomly initialize $\mathbf{g}^{(0)}=(g_1^{(0)},\dots,g_C^{(0)})$, where $g_i^{(0)}\in\mathcal{G}$}

\BlankLine
\For{each class $i\in\mathcal{C}$}{
    Compute reference centroid:
    $\displaystyle r_i \leftarrow \Big\|\frac{1}{K_i}\sum_{x\in\mathcal{D}_i}\phi(x)\Big\|_2$\;
    Generate $G$ subgroups $S_{i,g_i}$ with $K$ samples using $\mathcal{G}_{\theta'}$\;
    \For{$g_i=1$ \emph{to} $G$}{
        Compute subgroup centroid: $\displaystyle c_{i,g_i}\leftarrow \Big\|\frac{1}{K}\sum_{x\in S_{i,g_i}}\phi(x)\Big\|_2$\;
    }
}

\BlankLine
Initialize $\mathbf{g}\leftarrow \mathbf{g}^{(0)}$\;
\While{not converged}{
    Update $\mathbf{g}$ to reduce $\mathcal{L}_{\mathrm{S^3}}(\mathbf{g})$ via Eq.~\eqref{eq:selection-obj}
}
Set $S_i^* \leftarrow S_{i,g_i^*}$ for all $i\in\mathcal{C}$ using $\mathbf{g}^*=\arg\min_{\mathbf{g}}\mathcal{L}_{\mathrm{S^3}}(\mathbf{g})$\;
\Return{$\mathcal{D}_s=\{S_i^*\}_{i=1}^{C}$}\;
\end{algorithm}

\noindent\textbf{Selection Objective.}
We select one subgroup per class by minimizing a objective that balances proximity to real centroids and inter-class separation:
\begin{equation}
\begin{aligned}
\label{eq:selection-obj}
\mathcal{L}_{\mathrm{S^3}}(\mathbf{g})
&=
\alpha
\sum_{i=1}^{C}
\log\!\big(1-\sigma(c_{i,g_i},\,r_i)\big)\\
&
-\frac{\beta}{(C-1)G}
\sum_{i=1}^{C}
\sum_{\substack{j=1\\ j\neq i}}^{C}
\sum_{g=1}^{G}
\log\!\big(1-\sigma(c_{i,g_i},\,c_{j,g})\big),
\end{aligned}
\end{equation}
where $\alpha,~\beta>0$ weight the two criteria~(analysis see \cref{subsec:ab}), and 
$\mathbf{g}=(g_1,\dots,g_C)$ is a vector of subgroup indices,
with $g_i \in \mathcal{G}$ indicating the selected subgroup for class $i$.
The first term contracts subgroup centroids toward their real counterparts, constraining synthetic data within the true distribution; 
the second term enlarges pairwise angles between subgroup centroids, preserving diversity and mitigating mode collapse.

Because \cref{eq:selection-obj} depends only on precomputed centroids, 
we obtain the optimal subset of indices via
\begin{equation}
\label{eq:s3-opt}
\mathbf{\mathbf{g}}^*
=\mathop{\arg\min}_{\mathbf{g}}
\mathcal{L}_{\mathrm{S^3}}(\mathbf{g}),
\end{equation}
where $\mathbf{\mathbf{g}}^*=\{(i,g_i^*)\}_{i=1}^{C}$ denotes the indices of the optimal subgroup indices. \cref{eq:s3-opt} is solved using a simple greedy search strategy.
These indices are then used to find the corresponding subgroups and construct the distilled dataset  $\mathcal{D}_s$.

\cref{alg:s3} describes the sampling stage, which selects one representative subgroup per class by balancing proximity to real centroids and separation from other classes. Overall, this sampling strategy efficiently enhances the discriminative structure of the distilled dataset while preserving its representativeness and diversity.

\sectionnotoc{Experiments}
\label{sec:exps}

\begin{table*}[t]
\centering
\caption{Comparison of SOTA dataset distillation methods on ImageWoof under various IPC and backbone settings. The best results are marked as bold, and the second are underlined.}
\resizebox{\linewidth}{!}{
\begin{tabular}{cc|cccccccccc|c|c}
\toprule 
IPC (Ratio) & Model & Random & K-Center~\cite{k-center} & Herding~\cite{herding} & DiT~\cite{DiT} & Minimax~\cite{minimax} & DM~\cite{DM} & IDC-1~\cite{IDC} & D$^4$M~\cite{D4M} & D3HR~\cite{D3HR} & DDVLCP~\cite{ddvlcp} & \ims~(Ours) & Full \\
\midrule 
\multirow{3}{*}{10~(0.8\%)} 
& ConvNet-6   & 24.3{\tiny$\pm$1.1} & 19.4{\tiny$\pm$0.9} & 26.7{\tiny$\pm$0.5} & 34.2{\tiny$\pm$1.1} & 34.1{\tiny$\pm$0.4} & 26.9{\tiny$\pm$1.2} & 33.3{\tiny$\pm$1.1} & 29.4{\tiny$\pm$0.9} & -- & \underline{34.8{\tiny$\pm$2.4}} & \textbf{35.6{\tiny$\pm$1.8}} & 86.4{\tiny$\pm$0.2} \\
& ResNetAP-10 & 29.4{\tiny$\pm$0.8} & 22.1{\tiny$\pm$0.1} & 32.0{\tiny$\pm$0.3} & 34.7{\tiny$\pm$0.5} & 35.7{\tiny$\pm$0.3} & 30.3{\tiny$\pm$1.2} & 39.1{\tiny$\pm$0.5} & 33.2{\tiny$\pm$2.1} & \underline{40.7{\tiny$\pm$1.0}} & 39.5{\tiny$\pm$1.5} & \textbf{41.8{\tiny$\pm$0.3}} & 87.5{\tiny$\pm$0.5} \\
& ResNet-18   & 27.7{\tiny$\pm$0.9} & 21.1{\tiny$\pm$0.4} & 30.2{\tiny$\pm$1.2} & 34.7{\tiny$\pm$0.4} & 37.6{\tiny$\pm$0.9} & 33.4{\tiny$\pm$0.7} & 37.3{\tiny$\pm$0.2} & 32.3{\tiny$\pm$1.2} & 39.6{\tiny$\pm$1.0} & \underline{39.9{\tiny$\pm$2.6}} & \textbf{41.3{\tiny$\pm$1.1}} & 89.3{\tiny$\pm$1.2} \\
\midrule
\multirow{3}{*}{20~(1.6\%)} 
& ConvNet-6   & 29.1{\tiny$\pm$0.7} & 21.5{\tiny$\pm$0.8} & 29.5{\tiny$\pm$0.3} & 36.1{\tiny$\pm$0.8} & 37.9{\tiny$\pm$0.2} & 29.9{\tiny$\pm$1.0} & 35.5{\tiny$\pm$0.8} & 34.0{\tiny$\pm$2.3} & -- & \underline{37.9{\tiny$\pm$1.9}} & \textbf{39.4{\tiny$\pm$0.5}} & 86.4{\tiny$\pm$0.2} \\
& ResNetAP-10 & 32.7{\tiny$\pm$0.4} & 25.1{\tiny$\pm$0.7} & 34.9{\tiny$\pm$0.1} & 41.1{\tiny$\pm$0.8} & 43.3{\tiny$\pm$0.3} & 35.2{\tiny$\pm$0.6} & 43.4{\tiny$\pm$0.3} & 40.1{\tiny$\pm$1.6} & -- & \underline{44.5{\tiny$\pm$2.2}} & \textbf{45.8{\tiny$\pm$1.2}} & 87.5{\tiny$\pm$0.5} \\
& ResNet-18   & 29.7{\tiny$\pm$0.5} & 23.6{\tiny$\pm$0.3} & 32.2{\tiny$\pm$0.6} & 40.5{\tiny$\pm$0.5} & 40.9{\tiny$\pm$0.6} & 29.8{\tiny$\pm$1.7} & 38.6{\tiny$\pm$0.2} & 38.4{\tiny$\pm$1.1} & -- & \underline{44.5{\tiny$\pm$2.0}} & \textbf{44.9{\tiny$\pm$1.6}} & 89.3{\tiny$\pm$1.2} \\
\midrule
\multirow{3}{*}{50~(3.8\%)} 
& ConvNet-6   & 41.3{\tiny$\pm$0.6} & 36.5{\tiny$\pm$1.0} & 40.3{\tiny$\pm$0.7} & 46.5{\tiny$\pm$0.8} & 51.4{\tiny$\pm$0.4} & 44.4{\tiny$\pm$1.0} & 43.9{\tiny$\pm$1.2} & 47.4{\tiny$\pm$0.9} & -- & \underline{54.5{\tiny$\pm$0.6}} & \textbf{54.6{\tiny$\pm$0.1}} & 86.4{\tiny$\pm$0.2} \\
& ResNetAP-10 & 47.2{\tiny$\pm$1.3} & 40.6{\tiny$\pm$0.4} & 49.1{\tiny$\pm$0.7} & 49.3{\tiny$\pm$0.2} & 54.4{\tiny$\pm$0.6} & 47.1{\tiny$\pm$1.1} & 48.3{\tiny$\pm$1.0} & 51.7{\tiny$\pm$3.2} & \underline{59.3{\tiny$\pm$0.4}} & 57.3{\tiny$\pm$0.5} & \textbf{60.1{\tiny$\pm$0.7}} & 87.5{\tiny$\pm$0.5} \\
& ResNet-18   & 47.9{\tiny$\pm$1.8} & 39.6{\tiny$\pm$1.0} & 48.3{\tiny$\pm$1.2} & 50.1{\tiny$\pm$0.5} & 53.9{\tiny$\pm$0.6} & 46.2{\tiny$\pm$0.6} & 48.3{\tiny$\pm$0.8} & 53.7{\tiny$\pm$2.2} & 57.6{\tiny$\pm$0.4} & \underline{58.9{\tiny$\pm$1.5}} & \textbf{60.1{\tiny$\pm$1.1}} & 89.3{\tiny$\pm$1.2} \\
\midrule
\multirow{3}{*}{70~(5.4\%)} 
& ConvNet-6   & 46.3{\tiny$\pm$0.6} & 38.6{\tiny$\pm$0.7} & 46.2{\tiny$\pm$0.6} & 50.1{\tiny$\pm$1.2} & 51.8{\tiny$\pm$0.6} & 47.5{\tiny$\pm$0.8} & 48.9{\tiny$\pm$0.7} & 50.5{\tiny$\pm$0.4} & -- & \underline{55.8{\tiny$\pm$1.7}} & \textbf{56.0{\tiny$\pm$0.3}} & 86.4{\tiny$\pm$0.2} \\
& ResNetAP-10 & 50.8{\tiny$\pm$0.6} & 45.9{\tiny$\pm$1.5} & 53.4{\tiny$\pm$1.4} & 54.3{\tiny$\pm$0.9} & 56.5{\tiny$\pm$0.5} & 51.7{\tiny$\pm$0.8} & 52.8{\tiny$\pm$1.8} & 54.7{\tiny$\pm$1.6} & -- & \underline{60.6{\tiny$\pm$0.3}} & \textbf{61.0{\tiny$\pm$0.3}} & 87.5{\tiny$\pm$0.5} \\
& ResNet-18   & 52.1{\tiny$\pm$1.0} & 44.6{\tiny$\pm$1.1} & 49.7{\tiny$\pm$0.8} & 51.5{\tiny$\pm$1.0} & 52.7{\tiny$\pm$0.4} & 51.9{\tiny$\pm$0.8} & 51.1{\tiny$\pm$1.7} & 56.3{\tiny$\pm$1.8} & -- & \underline{60.3{\tiny$\pm$0.3}} & \textbf{61.1{\tiny$\pm$0.7}} & 89.3{\tiny$\pm$1.2} \\
\midrule
\multirow{3}{*}{100~(7.7\%)} 
& ConvNet-6   & 52.2{\tiny$\pm$0.4} & 45.1{\tiny$\pm$0.5} & 54.4{\tiny$\pm$1.1} & 53.4{\tiny$\pm$0.3} & 55.5{\tiny$\pm$0.6} & 55.0{\tiny$\pm$1.3} & 53.2{\tiny$\pm$0.9} & 57.9{\tiny$\pm$1.5} & -- & \textbf{62.7{\tiny$\pm$1.4}} & \underline{61.0{\tiny$\pm$0.6}} & 86.4{\tiny$\pm$0.2} \\
& ResNetAP-10 & 59.4{\tiny$\pm$1.0} & 54.8{\tiny$\pm$0.2} & 61.7{\tiny$\pm$0.9} & 58.3{\tiny$\pm$0.8} & 59.9{\tiny$\pm$0.4} & 56.4{\tiny$\pm$0.8} & 56.1{\tiny$\pm$0.9} & 59.5{\tiny$\pm$1.8} & 64.7{\tiny$\pm$0.3} & \underline{65.7{\tiny$\pm$0.5}} & \textbf{66.0{\tiny$\pm$0.8}} & 87.5{\tiny$\pm$0.5} \\
& ResNet-18   & 61.5{\tiny$\pm$1.3} & 50.4{\tiny$\pm$0.4} & 59.3{\tiny$\pm$0.7} & 58.9{\tiny$\pm$1.3} & 57.6{\tiny$\pm$0.6} & 60.2{\tiny$\pm$1.0} & 58.3{\tiny$\pm$1.2} & 63.8{\tiny$\pm$1.3} & 66.8{\tiny$\pm$0.6} & \underline{68.3{\tiny$\pm$0.4}} & \textbf{68.4{\tiny$\pm$0.8}} & 89.3{\tiny$\pm$1.2} \\
\bottomrule
\end{tabular}}
\label{tab:imagewoof}
\end{table*}

\begin{table}[t]
\centering
\caption{Comparison on ImageNette under different IPC settings. All the results are obtained on ResNetAP-10. The best results are marked as bold and the second are underlined.}\label{Nette}
\resizebox{\linewidth}{!}{
\begin{tabular}{ccccc|c}
\toprule
IPC & Random & DiT~\cite{DiT} & Minimax~\cite{minimax} & D$^4$M~\cite{D4M} & \ims~(Ours) \\
\midrule
10 & 54.2{\tiny$\pm$1.6} & 59.1{\tiny$\pm$0.7} & 58.6{\tiny$\pm$0.4} & \underline{60.9{\tiny$\pm$1.7}} & \textbf{62.9{\tiny$\pm$1.2}} \\
20 & 63.5{\tiny$\pm$0.5} & 64.8{\tiny$\pm$1.2} & \underline{70.6{\tiny$\pm$0.3}} & 66.3{\tiny$\pm$1.3} & \textbf{71.6{\tiny$\pm$0.7}} \\
50 & 76.1{\tiny$\pm$1.1} & 73.3{\tiny$\pm$0.9} & \underline{83.7{\tiny$\pm$0.4}} & 77.7{\tiny$\pm$1.1} & \textbf{84.2{\tiny$\pm$1.0}} \\
\bottomrule
\end{tabular}}
\end{table}

\begin{table}[t]
\centering
\caption{Comparison on ImageIDC under different IPC settings. All the results are obtained on ResNet-18. The best results are marked as bold, and the second are underlined.}\label{tab:idc}
\resizebox{\linewidth}{!}{
\begin{tabular}{cccccc|c}
\toprule
IPC & Random & SRe$^2$L~\cite{Sre2L}& DiT~\cite{DiT} &  RDED~\cite{RDED} &Minimax~\cite{IDC} & \ims~(Ours) \\
\midrule
1  & 12.7{\tiny$\pm$0.5} &17.2{\tiny$\pm$1.2} & \underline{26.7{\tiny$\pm$1.4}} &  15.9{\tiny$\pm$0.9} &22.4{\tiny$\pm$0.6} & \textbf{28.5{\tiny$\pm$2.0}} \\
10 & 48.1{\tiny$\pm$0.8} &23.6{\tiny$\pm$2.0}& \underline{54.1{\tiny$\pm$0.4}} &  33.8{\tiny$\pm$2.0} &51.9{\tiny$\pm$1.4} & \textbf{56.2{\tiny$\pm$0.6}} \\
50 & 68.1{\tiny$\pm$0.7} &45.7 {\tiny$\pm$1.3}& 76.4{\tiny$\pm$0.3} &  \underline{78.3{\tiny$\pm$0.9}} &78.3{\tiny$\pm$0.2} & \textbf{78.5{\tiny$\pm$1.5}} \\
\bottomrule
\end{tabular}}
\end{table}

\subsectionnotoc{Experimental Settings}
\label{subsec:expset}

\noindent\textbf{Datasets.}
We evaluate our method on high-resolution benchmarks to assess its scalability and generalization.
Experiments are conducted on ImageNet~\cite{imagenet15russakovsky} and it's subsets, including ImageNette~\cite{Howard_Imagenette_2019}, ImageWoof~\cite{Howard_Imagewoof_2019}, and ImageIDC\cite{IDC} with 256$\times$256 resolution, as well as ImageNet-100~\cite{imagenet100}.
ImageNette consists of 10 categories each, with ImageWoof containing 10 fine-grained dog breeds that exhibit high inter-class similarity, making it a challenging benchmark.

\noindent\textbf{Implementation Details.}
We apply Difffit~\cite{difffit} for parameter-efficient fine-tuning on a pretrained DiT-XL/2~\cite{DiT}.  
All experiments in \cref{subsec: cmp-exp} are under different \textit{Images Per Class} (IPC) settings. 
For ImageWoof, we evaluate all IPC levels of 10, 20, 50, 70, and 100. 
For ImageNette, we report results under IPC of 10, 20, and 50. 
For ImageIDC, we conduct experiments with IPC of 1, 10, and 50.
The balancing coefficient $\lambda_{\mathrm{IM}}$ in \cref{loss} is set to 0.002. Both fine-tuning and sample generation are conducted at an image resolution of $256\times256$. The diffusion model is fine-tuned for 8 epochs with a mini-batch size of 8 using AdamW with a learning rate of $1\times10^{-3}$. During the~\sss, the subgroup number $G$ and the coefficients $\alpha$ and $\beta$ are selected per dataset.  
We also examine the effect of $K_i$, the number of real samples used to compute each reference centroid. Representative findings on ImageWoof are reported in \cref{subsec: analy}. 

All experiments are performed on a single NVIDIA A100 GPU with 40GB of memory. Additional implementation details and extended experimental results are provided in the Appendix.

\noindent\textbf{Baseline and Evaluation Metrics.}
We compare our method with both optimization-based and diffusion-based dataset distillation baselines, including K-Center~\cite{k-center}, Herding~\cite{herding}, DM~\cite{DM}, ~\cite{IDC}, SRe$^2$L~\cite{Sre2L}, RDED~\cite{RDED}, Minimax~\cite{minimax}, D$^4$M~\cite{D4M},, D$^3$HR~\cite{D3HR}, and DDVLCP~\cite{ddvlcp}.
Each distilled dataset is used to train classification models from scratch (\eg, ConvNet-6, ResNetAP-10, and ResNet-18) for 300 epochs with SGD and cosine learning rate decay, following the evaluation protocol introduced in Minimax~\cite{minimax}.
We report the mean and standard deviation of Top-1 Accuracy over three random seeds as the primary evaluation metric.
Currently, there are two evaluation paradigms for assessing distilled datasets: one based on hard labels~\cite{minimax}, where models are trained directly with one-hot supervision, and the other on soft labels~\cite{soft}, where knowledge distillation transfers teacher predictions as continuous targets.
Following the settings in~\cite{cao2}, we evaluate our method under both paradigms and report the higher accuracy between the two for each experiment.

\subsectionnotoc{Comparison Experiments}
\label{subsec: cmp-exp}

In this section, we present comprehensive comparisons between our approach and multiple state-of-the-art dataset distillation methods across three benchmark datasets: ImageWoof, ImageNette, and ImageIDC. Additional experimental results, including extended quantitative studies, qualitative visualizations, and evaluations on ImageNet-100 and ImageNet-1K, are provided in the Appendix.

\noindent\textbf{ImageWoof.}
We evaluate our method on the fine-grained ImageWoof dataset, which contains 10 dog breeds with high inter-class similarity.
\cref{tab:imagewoof} presents the comparison with representative optimization-based methods (\eg, SRe$^2$L~\cite{Sre2L}) and diffusion-based approaches including Minimax~\cite{minimax}, D$^4$M~\cite{D4M}, D$^3$HR~\cite{D3HR}, and DDVLCP~\cite{ddvlcp}.
The gains are especially pronounced under low IPC budgets.
For example, at IPC = 10,~\ims~improves ResNetAP-10 accuracy from 35.7\% achieved by Minimax~\cite{minimax} and 39.5\% achieved by DDVLCP~\cite{ddvlcp} to 41.8\%, corresponding to absolute improvements of 6.1\% and 2.3\% respectively.
On ResNet-18, the accuracy increases from 37.6\% with Minimax~\cite{minimax} and 39.9\% with DDVLCP~\cite{ddvlcp} to 41.3\%, yielding gains of 3.7\% and 1.4\%. These improvements indicate that~\ft~expands the model’s coverage of low-density yet discriminative regions, while \sss~enhances class separability by selecting well-aligned subgroups.
As IPC increases,~\ims~consistently superior, demonstrating robust performance across different compression ratios and confirming that our method yields synthetic datasets that are both representative and discriminatively structured.

\noindent\textbf{ImageNette.}
We evaluate our method on the high-resolution ImageNette dataset, which contains 10 natural object categories and provides a clean yet challenging benchmark for assessing the quality of distilled samples.
Using the ResNetAP-10 backbone, the results in \cref{Nette} show that our method consistently outperforms all diffusion-based baselines.
In particular, at IPC = 50, our approach achieves 84.2\%, surpassing the strongest baseline by 0.5\%.

\noindent\textbf{ImageIDC.}
We further evaluate our method on the ImageIDC dataset, which contains classes with less similarity, using ResNet-18 as the backbone. As shown in \cref{tab:idc}, our method consistently achieves the best performance across all IPC settings.
The improvement is most notable at IPC = 1, where our approach reaches 28.5\%, outperforming the strongest baseline by 0.5\%, demonstrating clear advantages in modeling the underlying distribution under extremely limited data budgets.

\begin{figure}[t]
  \centering
    \includegraphics[width=0.97\linewidth]{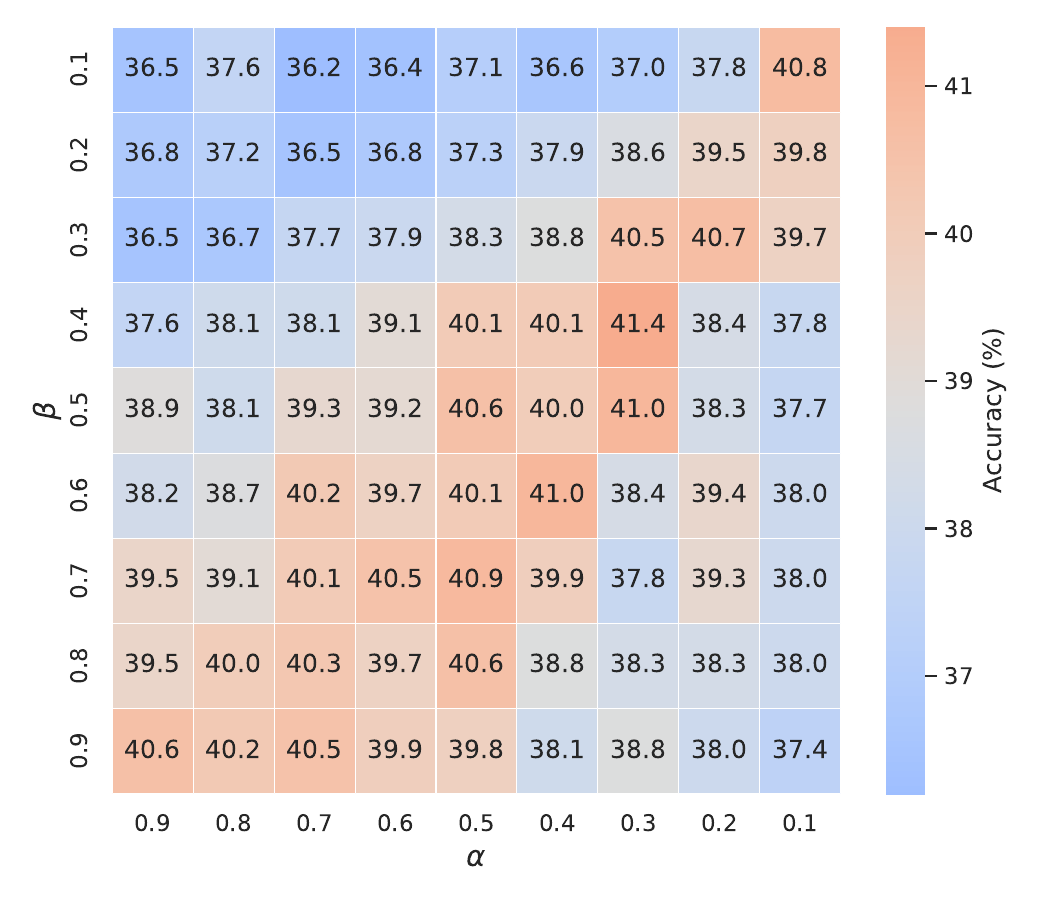}
  \caption{Heatmap of classification accuracy under different combinations of $\alpha$ and $\beta$.
        The heatmap shows that balanced values of $\alpha$ and $\beta$ yield the best performance.}
  \label{fig:heatmap_ab}
\end{figure}

\subsectionnotoc{Analysis Experiments}
\label{subsec: analy}

We analyze three key hyperparameters and conduct ablation studies for our proposed method. Additional analysis results, including analysis of $\lambda_{\mathrm{IM}}$ and $G$, see Appendix.

\noindent\textbf{Analysis on $\alpha$ and $\beta$.}\label{subsec:ab}
To further analyze the impact of the weighting coefficients in the subgroup selection objective, 
we conduct a sensitivity study on the parameters $\alpha$ and $\beta$ in \cref{eq:selection-obj}. 
\cref{fig:heatmap_ab} presents the accuracy heatmap obtained by varying $\alpha$ and $\beta$ within the range $[0.1, 0.9]$.
The results show that model performance remains stable across a wide range of parameter combinations, 
demonstrating the robustness of our method.

\noindent\textbf{Ablation Studies.}\label{subsec: abl}
\cref{tab:ablation} presents the ablation results of~\ft~and~\sss~on ImageWoof and ImageNette under IPC settings of 10 and 50.
Both components individually improve over the DiT baseline:~\ft~expands distributional coverage, while~\sss~strengthens inter-class separation during sampling.
When combined as~\ims, the improvements are consistently larger across all settings.

\begin{table}[t]
\centering
\caption{Ablation study of~\ft~and~\sss~ under IPC = 10 and 50. Results on ImageWoof are obtained using ResNetAP-10, while results on ImageNette are obtained using ResNet-18. The best results are marked as bold.}
\resizebox{\columnwidth}{!}{
\begin{tabular}{l|cc|cc}
\toprule
\multirow{2}{*}{Method} & 
\multicolumn{2}{c|}{ImageWoof} & 
\multicolumn{2}{c}{ImageNette} \\
\cmidrule(lr){2-3} \cmidrule(lr){4-5}
 & IPC = 10 & IPC = 50 & IPC = 10 & IPC = 50 \\
\midrule
DiT~\cite{DiT}         & 34.7$\pm$0.5       & 49.3$\pm$0.2       & 58.9$\pm$1.2  & 82.9$\pm$0.1 \\
~+ \ft       & 37.3 $\pm$0.1      & 53.5$\pm$0.5       & 60.0$\pm$0.9    & 81.5$\pm$0.5 \\
~+ \sss      & 40.9  $\pm$0.2     & 54.9 $\pm$0.5      & 62.2$\pm$1.9  & 81.3$\pm$1.2 \\
\rowcolor{gray!25}
~+ \ims      & \textbf{41.8 $\pm$0.3}      & \textbf{61.0 $\pm$0.7}      & \textbf{62.9$\pm$1.2}  & \textbf{84.2$\pm$1.0} \\
\bottomrule
\end{tabular}}
\label{tab:ablation}
\end{table}

\begin{figure}[t]
  \centering
    \includegraphics[width=0.95\linewidth]{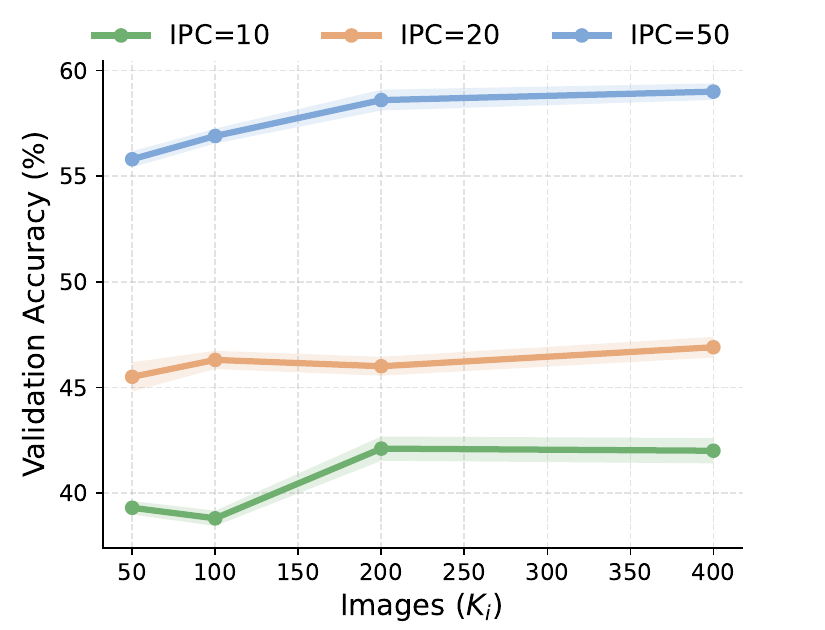}
  \caption{
    Analysis of real-images usage from original dataset used to compute reference centroids in \cref{eq:real-centroid}.
     Moderate sample counts provide stable and reliable centroids for subgroup selection, while too few samples introduce noise and degrade selection quality. }
  \label{fig:group_max}
\end{figure}

\noindent\textbf{Analysis on $K_i$.}\label{subsec:k}
\cref{fig:group_max} examines the effect of the real-image budget $K_i$ in \cref{eq:real-centroid} within our proposed~\sss.
We conduct experiments on ResNetAP-10 under IPC = 10, 20, and 50. The result demonstrates that use more images to compute centroids stabilizes the reference distribution and consistently improves the quality of the selected subgroups, especially under high IPC setting.

\sectionnotoc{Conclusion}
\label{sec:conclusion}
We propose \ims, a simple yet effective two-stage framework for large-scale dataset distillation, addressing two critical issues in diffusion-based approaches: insufficient distribution coverage and weak discriminative diversity. Our method achieves state-of-the-art performance across multiple benchmarks, including ImageWoof, ImageNette, ImageIDC, and ImageNet, under various evaluation settings. By combining inversion-guided fine-tuning with centroid-based sampling,~\ims~provides a new perspective and a practical solution for improving diffusion-based dataset distillation.

\section*{Acknowledgement}

This work was supported by the National Natural Science Foundation of China (No. 6250070674) and the Zhejiang Leading Innovative and Entrepreneur Team Introduction Program (2024R01007). This work was also supported by the National Research Foundation, Singapore and Infocomm Media Development Authority under its Trust Tech Funding Initiative. Any opinions, findings and conclusions or recommendations expressed in this material are those of the author(s) and do not reflect the views of the National Research Foundation, Singapore and Infocomm Media Development Authority.

{
    \small
    \bibliographystyle{ieeenat_fullname}
    \bibliography{main}
}

\clearpage
\setcounter{tocdepth}{0} 
\setcounter{page}{1}
\maketitlesupplementary

\setcounter{tocdepth}{2} 
\tableofcontents

\section{Instability in Inversion}
\label{sec:instability}

Diffusion inversion theoretically provides a deterministic mapping between real data and their latent noises by integrating the probability flow ODE (PF-ODE) backward in time. 
However, recent work reveals that this inversion process is inherently unstable in high-dimensional spaces~\cite{instability}. 
Specifically, even infinitesimal perturbations in the latent space can be amplified through the ODE dynamics, leading to substantial reconstruction errors during the forward regeneration process.

Formally, let $F:\mathbb{R}^n\!\rightarrow\!\mathbb{R}^n$ denote the PF-ODE mapping from noise to data. 
The instability of $F$ can be characterized by the geometric mean of its singular values, known as the \emph{geometric mean instability coefficient}:
\begin{equation}\label{eq:insta-coef}
\bar{\mathcal{E}}_F(\mathbf{z}) = 
\left(\prod_{i=1}^{n} 
\frac{\|J_F(\mathbf{z})\,\mathbf{u}_i\|_2}{\|\mathbf{u}_i\|_2}
\right)^{\!1/n},
\end{equation}
where $J_F(\mathbf{z})$ denotes the Jacobian matrix of $F$ at $\mathbf{z}$ and $\{\mathbf{u}_i\}$ are orthonormal basis vectors. 
When $\bar{\mathcal{E}}_F(\mathbf{z}) > 1$, the mapping locally expands the volume element around $\mathbf{z}$, implying that infinitesimal perturbations will be amplified after propagation through $F$.

To quantify the probability that such instability occurs, they establish the following lower bound for any threshold $M>1$:
\begin{equation}\label{eq:insta-prob}
\mathcal{P}_M := 
\pi_{\text{real}}\!\left(
\left\{
\mathbf{z} :
\bar{\mathcal{E}}_F(F^{-1}(\mathbf{z})) > M
\right\}
\right)
\ge 1 - \epsilon - \delta,
\end{equation}
where
\begin{equation}\label{eq:insta-terms}
\begin{aligned}
\epsilon &:= 
\pi_{\text{real}}\!\left(
\!\left\{
\mathbf{z} : 
p_{\text{gen}}(\mathbf{z}) 
\ge 
\frac{1}{(2\pi M^2)^{n/2}}
e^{-\frac{2n+3\sqrt{2n}}{2}}
\right\}
\!\right),\\
\delta &:= 
\pi_{\text{real}}\!\left(
\!\left\{
\mathbf{z} : 
\|F^{-1}(\mathbf{z})\|^2 
> 2n + 3\sqrt{2n}
\right\}
\!\right).
\end{aligned}
\end{equation}
Here, $\pi_{\text{real}}$ denotes the real data distribution, and $p_{\text{gen}}$ represents the generation distribution of the diffusion model. 
As dimensionality $n$ increases, both $\epsilon$ and $\delta$ vanish, implying that the instability probability $\mathcal{P}_M$ approaches one—thus instability becomes almost inevitable in high-dimensional diffusion mappings.

Intuitively, this phenomenon arises from the sparsity of the generation distribution. 
Since the generative probability mass is concentrated in a few narrow regions while most of the space exhibits near-zero density, the PF-ODE must stretch the mapping to preserve probability. 
Consequently, regions of low density in the generative space correspond to large local Jacobian norms, leading to the amplification of any perturbation during inversion or regeneration. 
This explains why inversion trajectories often drift toward low-density, high-sensitivity regions of the data manifold, as also observed in Sec.~\ref{sec:im}. 
A more detailed mathematical derivation and discussion of the instability mechanism can be found in ~\cite{instability}.

This inherent instability directly motivates the design of our Inversion-Matching fine-tuning (\ft) in ImS$^3$. 
Rather than viewing instability as an obstacle, we exploit it as a guiding property to drive the diffusion model toward low-density regions. 

By aligning the denoising trajectory with the inversion trajectory, our method explicitly leverages the instability-induced drift to expand the distributional coverage and enhance representation diversity in the distilled dataset.

\section{More Experiments}
\label{sec:moreexp}

\subsection{Comparison on ImageNet-100}
\label{subsec:image100}
\begin{table*}[t]
\centering
\caption{Performance comparison on ImageNet-100 under different IPC and backbone settings. 
Results are Top-1 accuracy.}
\resizebox{\linewidth}{!}{
\begin{tabular}{cc|ccccc|c}
\toprule
IPC & Model & Random & Herding~\cite{herding} & IDC-1~\cite{IDC} & Minimax~\cite{IDC} & MGD$^3$~\cite{mgd3} & \ims~(Ours) \\
\midrule
\multirow{3}{*}{10}
& ConvNet-6 
    & 17.0{\tiny$\pm$0.3} 
    & 17.2{\tiny$\pm$0.3} 
    & 24.3{\tiny$\pm$0.5} 
    & 22.3{\tiny$\pm$0.5} 
    & 23.4{\tiny$\pm$0.9} 
    & \textbf{24.3{\tiny$\pm$1.1}} \\
& ResNetAP-10 
    & 19.1{\tiny$\pm$0.4} 
    & 19.8{\tiny$\pm$0.3} 
    & 25.7{\tiny$\pm$0.1} 
    & 24.8{\tiny$\pm$0.2} 
    & 25.8{\tiny$\pm$0.5} 
    & \textbf{28.4{\tiny$\pm$0.2}} \\
& ResNet-18 
    & 17.5{\tiny$\pm$0.5} 
    & 16.1{\tiny$\pm$0.2} 
    & 25.1{\tiny$\pm$0.2} 
    & 22.5{\tiny$\pm$0.3} 
    & 23.6{\tiny$\pm$0.4} 
    & \textbf{28.3{\tiny$\pm$0.3}} \\
\midrule
\multirow{3}{*}{20}
& ConvNet-6
    & 24.8{\tiny$\pm$0.2}
    & 24.3{\tiny$\pm$0.4}
    & 28.8{\tiny$\pm$0.3}
    & 29.3{\tiny$\pm$0.4}
    & 30.6{\tiny$\pm$0.4}
    & \textbf{30.7{\tiny$\pm$0.4}} \\
& ResNetAP-10
    & 26.7{\tiny$\pm$0.5}
    & 27.6{\tiny$\pm$0.1}
    & 29.9{\tiny$\pm$0.2}
    & 32.3{\tiny$\pm$0.1}
    & 33.9{\tiny$\pm$1.1}
    & \textbf{34.8{\tiny$\pm$0.2}} \\
& ResNet-18
    & 25.5{\tiny$\pm$0.3}
    & 24.7{\tiny$\pm$0.1}
    & 30.2{\tiny$\pm$0.2}
    & 31.2{\tiny$\pm$0.1}
    & 32.6{\tiny$\pm$0.4}
    & \textbf{36.3{\tiny$\pm$0.2}} \\
\bottomrule
\end{tabular}\label{imagenet100}
}
\end{table*}

\cref{imagenet100} reports the comparison across different IPC and backbone on ImageNet-100. Across all settings, our method consistently delivers the highest accuracy, outperforming existing approaches such as Herding~\cite{herding}, IDC-1~\cite{IDC}, Minimax~\cite{minimax}, and MGD$^3$~\cite{mgd3}. 

\subsection{Experiment Results on Different Student Model}
\begin{table*}[t]
\centering
\caption{Performance comparison under different IPC and backbone settings on ImageWoof. Best results in each setting are marked as bold.}
\resizebox{\textwidth}{!}{
\begin{tabular}{c|c|cccc|c}
\toprule
Backbone & IPC
& SRe$^2$L~\cite{Sre2L}
& Minimax~\cite{minimax}
& RDED~\cite{RDED}
& CaO$^2$~\cite{cao2}
& \ims~(Ours) \\
\midrule

\multirow{2}{*}{ResNet-18}
& 10 & 20.2{\tiny$\pm$0.2} & 40.1{\tiny$\pm$1.0} & 38.5{\tiny$\pm$2.1} & 45.6{\tiny$\pm$1.4} & \textbf{45.9{\tiny$\pm$1.3}} \\
& 50 & 23.3{\tiny$\pm$0.3} & 67.0{\tiny$\pm$1.8} & 68.5{\tiny$\pm$0.7} & 68.9{\tiny$\pm$1.1} & \textbf{71.2{\tiny$\pm$1.3}} \\
\midrule

\multirow{2}{*}{ResNet-50}
& 10 & 17.3{\tiny$\pm$1.7} & 37.3{\tiny$\pm$1.1} & 29.9{\tiny$\pm$2.2} & 40.1{\tiny$\pm$0.1} & \textbf{42.7{\tiny$\pm$1.8}} \\
& 50 & 24.8{\tiny$\pm$0.7} & 64.3{\tiny$\pm$0.9} & 67.8{\tiny$\pm$0.3} & 68.2{\tiny$\pm$1.1} & \textbf{68.3{\tiny$\pm$0.2}} \\
\midrule

\multirow{2}{*}{ResNet-101}
& 10 & 17.7{\tiny$\pm$0.9} & 34.2{\tiny$\pm$1.7} & 31.3{\tiny$\pm$1.3} & 36.5{\tiny$\pm$1.4} & \textbf{38.2{\tiny$\pm$1.6}} \\
& 50 & 21.2{\tiny$\pm$0.2} & 62.7{\tiny$\pm$1.6} & 59.1{\tiny$\pm$0.7} & 63.1{\tiny$\pm$1.3} & \textbf{66.1{\tiny$\pm$1.8}} \\
\bottomrule
\end{tabular}\label{backbone}
}
\end{table*}

\cref{backbone} reports the performance of different dataset distillation methods on ImageWoof across multiple backbone architectures,including ResNet-18, ResNet-50, ResNet-101, and multiple IPC settings.
Across all configurations, our method achieves the highest accuracy, demonstrating clear and consistent advantages in both low- and high-budget regimes.
The improvements are particularly notable under low IPC (e.g., IPC = 10), where the limited data makes representation learning more challenging: \ims\ boosts ResNet-18 performance from 45.6\% (CaO$^2$) to 45.9\%, and ResNet-50 accuracy from 40.1\% to 42.7\%.
Under higher IPC (IPC = 50), \ims\ continues to outperform strong baselines such as Minimax~\cite{minimax}, RDED~\cite{RDED}, and CaO$^2$~\cite{cao2}, confirming that our approach generalizes well across model capacities and data regimes.
These results highlight the robustness and scalability of \ims\ for improving distilled dataset quality on large-scale visual recognition tasks.

\subsection{Analysis on Loss Functions}
\label{subsec:loss}
\begin{table}[t]
\centering
\caption{Ablation study of different similarity losses used in the~\ft~
on \textbf{ImageWoof} and \textbf{ImageNette} under IPC = 10 and 50. 
Results are reported as top-1 accuracy.}
\resizebox{\columnwidth}{!}{
\begin{tabular}{c|cc|cc}
\toprule
\multirow{2}{*}{Loss Type} & 
\multicolumn{2}{c|}{ImageWoof} & 
\multicolumn{2}{c}{ImageNette} \\
\cmidrule(lr){2-3} \cmidrule(lr){4-5}
 & IPC = 10 & IPC = 50 & IPC = 10 & IPC = 50 \\
\midrule
$L_1$  & 38.7{\tiny$\pm$0.2} & 58.1{\tiny$\pm$0.2} & 62.6{\tiny$\pm$0.9} & 81.7{\tiny$\pm$0.7} \\
$L_2$  & 39.0{\tiny$\pm$0.2} & 58.1{\tiny$\pm$0.1} & 62.7{\tiny$\pm$1.0} & 82.8{\tiny$\pm$0.7} \\
\rowcolor{gray!25}
$1-\sigma$ & \textbf{41.8{\tiny$\pm$0.3}} & \textbf{60.1{\tiny$\pm$0.7}} & \textbf{62.9{\tiny$\pm$1.2}} & \textbf{84.2{\tiny$\pm$1.0}} \\
\bottomrule
\end{tabular}}
\label{tab:loss_ablation}
\end{table}

Table~\ref{tab:loss_ablation} presents an ablation study on different similarity
losses used in~\ft. We evaluate three widely
used losses $L_1$, $L_2$, and $1-\sigma$ in~\cref{sec:im} on
ImageWoof and ImageNette under IPC = 10 and 50. Both $L_1$ and $L_2$ exhibit comparable
performance, providing moderate improvements across IPC settings. In contrast,
the proposed $1-\sigma$ formulation consistently achieves the highest accuracy,
surpassing the other two losses by a clear margin at both data budgets. These
results highlight the importance of treating feature deviations in a
distribution-aware manner, suggesting that the $1-\sigma$-based loss stabilizes
inversion alignment and leads to more reliable distilled samples.

\subsection{Analysis on $\lambda_{\text{IM}}$}
\label{subsec:abl_lambda}
\begin{figure}[t]
  \centering
    \includegraphics[width=\linewidth]{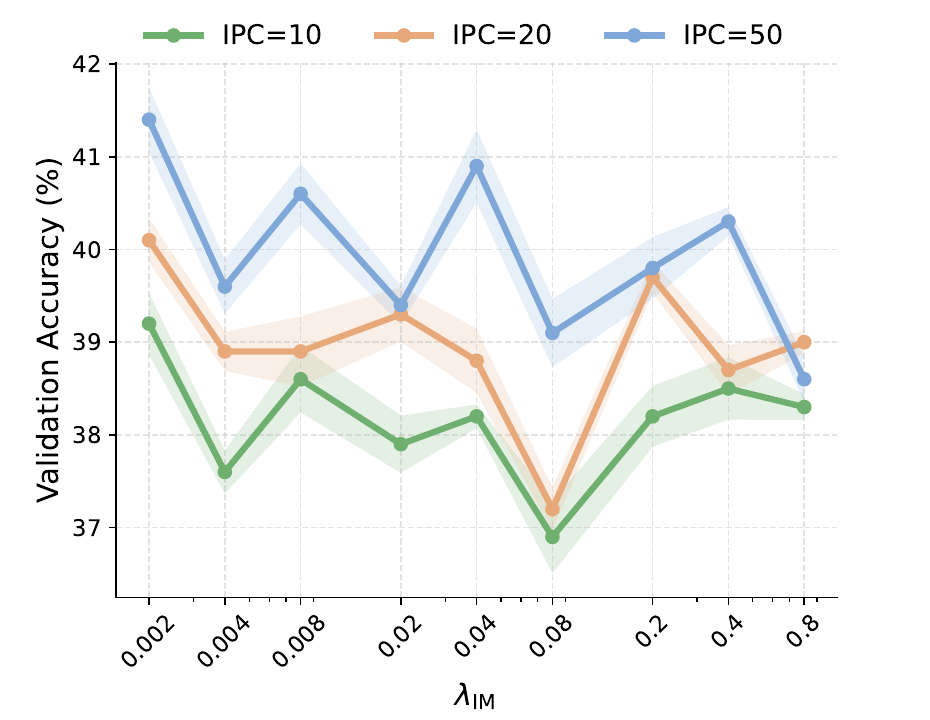}
    \caption{
        Validation accuracy under different matching strengths.
Performance of our method across a range of matching coefficients $\lambda_{\mathrm{IM}}$ for IPC = 10, 20, and 50. Each curve corresponds to a fixed IPC, and shaded regions denote performance variability across runs.
}
  \label{fig:lambda-sweep}
\end{figure}

To examine the sensitivity of our method to the matching strength, we perform 
a sweep over $\lambda_{\text{IM}}$ and report the validation accuracy for 
IPC = 10, 20, and 50 in \cref{fig:lambda-sweep}). 
Across all settings, the performance remains stable within a broad interval 
of $\lambda_{\text{IM}}$, indicating that the method is not overly 
sensitive to this hyperparameter. 
For smaller coefficients (\emph{e.g.,} $2\times10^{-3}$ to $8\times10^{-3}$), the 
accuracy stays consistently high, especially at IPC = 50. 
When $\lambda_{\text{IM}}$ becomes excessively large (\emph{e.g.,} 0.4--0.8), 
the accuracy gradually drops, likely because overly strong alignment introduces 
optimization bias and reduces the diversity of the distilled samples.

\subsection{Analysis on $G$}
\label{subsec:abl_k}
\begin{figure}[t]
  \centering
    \includegraphics[width=\linewidth]{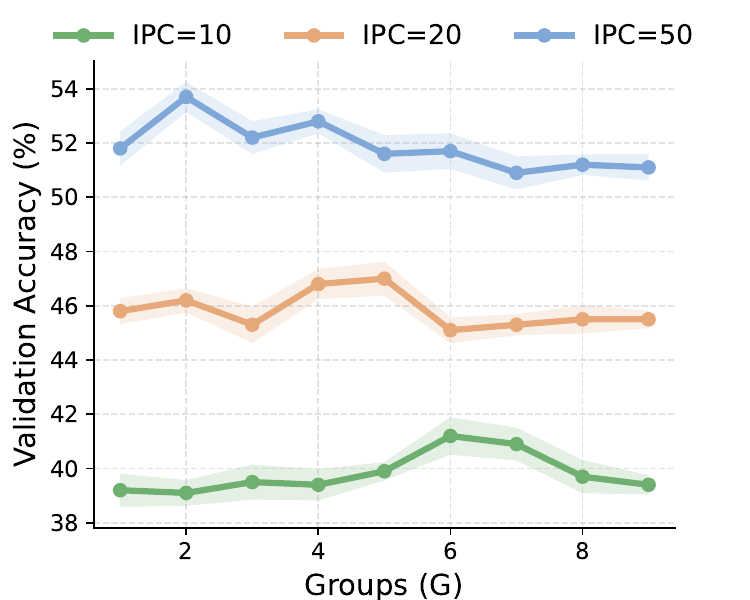}
    \caption{
        Effect of subgroup pool size $G$ under different IPC settings on ImageWoof. 
        A moderate increase in $G$ provides richer intra-class variation and improves selection quality, especially in low-IPC regimes. However, excessively large pools introduce redundant or noisy candidates, which destabilizes selection and leads to degraded performance.
}
  \label{fig:group}
\end{figure}

We analyze the effect of the subgroup pool size $G$, which determines how many candidate subgroups are generated for each class before selection. As shown in~\cref{fig:group}, increasing $G$ helps expose more inter-class variability and improves distilled accuracy when the IPC is small, since low-data regimes benefit from a richer candidate pool. However, the trend does not extend indefinitely: overly large pools introduce redundant or low-quality samples, weakening the discriminative structure and causing performance to decline. This highlights the need to choose $G$ according to the IPC setting, where lower IPC typically requires a larger pool while higher IPC benefits less from further expansion.

\subsection{Analysis on Feature Extractors for Centroid Selection}
\label{subsec:abl_feat}

To further investigate the robustness of our centroid–based subgroup selection,
we replace the default ResNet-18 feature extractor with different backbones, including ResNet-50, ResNet-101, EfficientNet, and CLIP.
We report the distilled top-1 accuracy on ImageWoof under IPC = 10 and 20.

\begin{table}[h!]
\centering
\caption{Ablation study on feature extractors for centroid selection on ImageWoof. 
Top-1 accuracy (\%).}
\label{tab:feat_ablation}
\resizebox{\linewidth}{!}{
\begin{tabular}{c|c|c|c|c|c}
\toprule
IPC & Res18 & Res50 & Res101 & Efficient & CLIP \\
\midrule
10 & \textbf{41.8{\tiny$\pm$0.3}} 
   & 39.6{\tiny$\pm$0.3} 
   & 38.8{\tiny$\pm$0.2} 
   & 39.5{\tiny$\pm$0.2} 
   & 38.1{\tiny$\pm$0.5} \\
20 & \textbf{45.8{\tiny$\pm$1.2}} 
   & 45.7{\tiny$\pm$0.3} 
   & 45.0{\tiny$\pm$0.2} 
   & 45.8{\tiny$\pm$0.3} 
   & 44.3{\tiny$\pm$0.2} \\
\bottomrule
\end{tabular}}
\end{table}
Across both settings, we observe that the choice of feature extractor has noticable impact on the quality of sampling process.
The results do not follow the conventional expectation that deeper backbones always yield better distilled results. Instead, the lightweight ResNet-18 achieves the highest top-1 accuracy among both IPC budgets. It implies that moderate capacity features provide a more stable and suitable
embedding space for centroid sampling procedure.

\subsection{t-SNE Visualization on ImageNet-100}
\label{subsec:tsne_imagenet100}

\begin{figure*}[t]
    \centering

    \begin{subfigure}{0.45\linewidth}
        \centering
        \includegraphics[width=\linewidth]{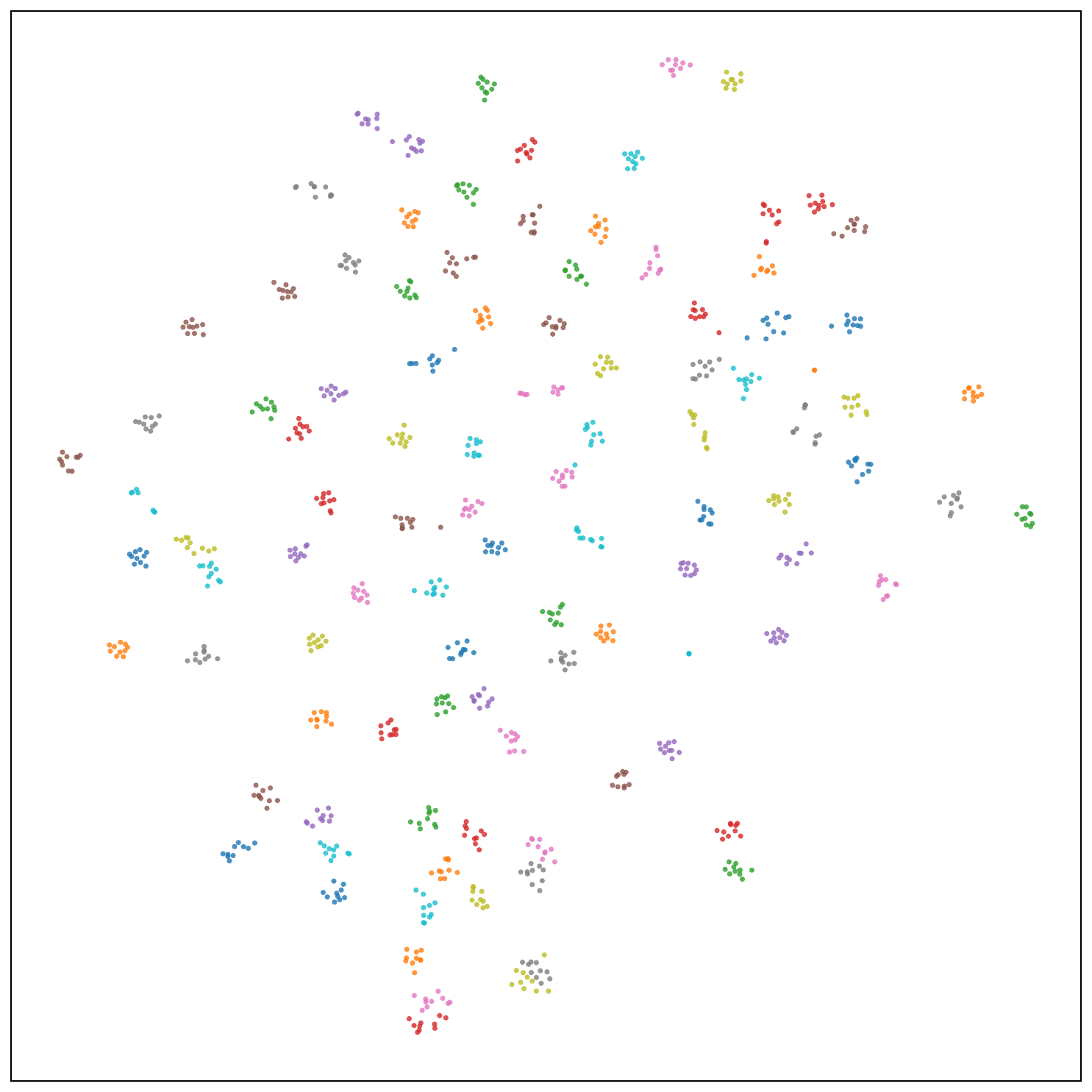}
        \caption{ImageNet-100: ~\ims~ (ours)}
    \end{subfigure}
    \hfill
    \begin{subfigure}{0.45\linewidth}
        \centering
        \includegraphics[width=\linewidth]{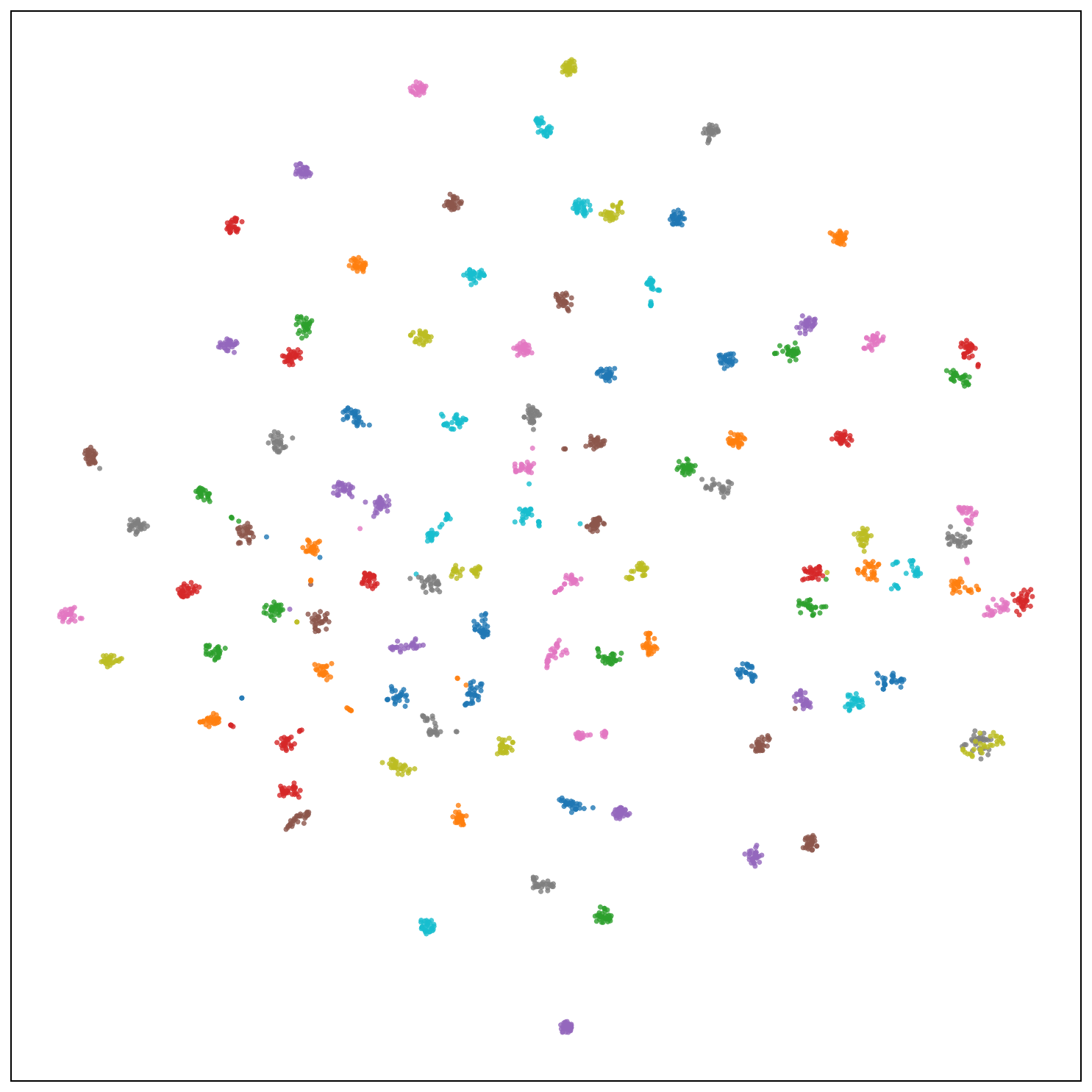}
        \caption{ImageNet-100: MinimaxDiffusion}
    \end{subfigure}

    \caption{
        t-SNE visualization of distilled samples on ImageNet-100 and ImageWoof.
        ~\ims~ produces tighter intra-class clusters and clearer inter-class separation,
        whereas MinimaxDiffusion exhibits more dispersed and overlapping structures.
    }
    \label{fig:tsne_all}
\end{figure*}

To further examine how our method shapes the feature distribution of the distilled dataset,
we visualize the embeddings of synthesized samples using t-SNE on ImageNet-100.
We compare our approach (\ims) with MinimaxDiffusion, and the results are shown in
\cref{fig:tsne_all}. Each color corresponds to a different class.

As illustrated by the comparison,~\ims~produces significantly more compact and
well-structured intra-class clusters. Points belonging to the same class exhibit
tighter grouping, indicating that our inversion-matching mechanism encourages
stronger within-class consistency. Moreover,~\ims~demonstrates clearer separation
between different classes, suggesting that the synthesized samples form a more
discriminative embedding structure. This indicates that our distribution-aware
feature alignment promotes both intra-class cohesion and inter-class
separability, which are crucial properties for effective dataset distillation.

In contrast, MinimaxDiffusion yields more fragmented and overlapping clusters.
Classes become less compact and boundaries are looser, implying weaker semantic
organization in the distilled data. The tighter clustering and improved class
separation produced by~\ims~provide intuitive evidence that our method captures
the intrinsic class geometry more faithfully, leading to improved downstream
recognition performance.

\subsection{Runtime Analysis}
We report runtime on a single RTX 4090 GPU using ImageWoof. As shown in \Cref{tab:runtime}, IM requires only 8 epochs of fine-tuning, while S$^3$ sampling is training-free, resulting in competitive total runtime.

\begin{table}[h]

\centering
\caption{Runtime comparison on ImageWoof (IPC=10).}
\label{tab:runtime}
\resizebox{\linewidth}{!}{
\begin{tabular}{l|c|ccc}
\toprule
 & DiT & Minimax & MGD3 & \textbf{IMS$^3$} \\
\midrule
Fine-tuning & - & 32min & - & \textbf{19.5min} \\
Sampling & 1.5min & \textbf{1.5min} & 21min & 7min \\
Total & 1.5min & 33.5min & \textbf{21min} & 26.5min \\
\bottomrule
\end{tabular}
}

\end{table}

\subsection{Higher Resolution (512$\times$512).}
We evaluate IMS$^3$ at 512$\times$512 using DiT-XL/2-512. \Cref{tab:512} reports consistent advantages at IPC=10 and IPC=50.

\begin{table}[h]

\centering
\caption{512$\times$512 results on ImageWoof.}
\label{tab:512}
\resizebox{\linewidth}{!}{
\begin{tabular}{lcccc}
\toprule
 & DiT & Minimax & CaO$^2$ & \textbf{IMS$^3$} \\
\midrule
IPC=10 & 36.6{\tiny$\pm$0.2} & 33.3{\tiny$\pm$0.2} & 37.5{\tiny$\pm$0.3} & \textbf{37.9{\tiny$\pm$0.2}} \\
IPC=50 & 52.5{\tiny$\pm$0.2} & 51.4{\tiny$\pm$0.5} & 53.8{\tiny$\pm$0.6} & \textbf{59.6{\tiny$\pm$0.3}} \\
\bottomrule
\end{tabular}
}

\end{table}

\subsection{ViT Architecture Evaluation.}
We evaluate distilled datasets on ViT and EfficientNet. As shown in \Cref{tab:vit}, IMS$^3$ improves results under both architectures, indicating good transfer across classifier families.

\begin{table}[h]

\centering
\caption{Cross-architecture evaluation on ImageWoof.}
\label{tab:vit}
\resizebox{\linewidth}{!}{
\begin{tabular}{lc|ccc}
\toprule
 & Model & Minimax & MGD3 & \textbf{IMS$^3$} \\
\midrule
\multirow{2}{*}{IPC=10}
& EfficientNet& 31.3{\tiny$\pm$0.7} & 32.3{\tiny$\pm$0.3} & \textbf{34.2{\tiny$\pm$0.2}} \\
& ViT          & 18.9{\tiny$\pm$0.0} & 17.7{\tiny$\pm$0.2} & \textbf{20.0{\tiny$\pm$0.0}} \\
\midrule
\multirow{2}{*}{IPC=50}
& EfficientNet & 46.3{\tiny$\pm$1.0} & 49.3{\tiny$\pm$0.5} & \textbf{51.7{\tiny$\pm$0.4}} \\
& ViT         & 18.8{\tiny$\pm$0.3} & 17.6{\tiny$\pm$0.1} & \textbf{19.9{\tiny$\pm$0.5}} \\
\bottomrule
\end{tabular}
}

\end{table}

\subsection{ImageNet-1K Experiments.}
We evaluate IMS$^3$ on full ImageNet-1K. \Cref{tab:imagenet1k} reports results.

\begin{table}[h]

\centering
\caption{ImageNet-1K results (IPC=10).}
\label{tab:imagenet1k}
\resizebox{\linewidth}{!}{
\begin{tabular}{lccccc}
\toprule
 & SRe$^2$L & RDED & Minimax & MGD3 & \textbf{IMS$^3$} \\
\midrule
Top-1 Acc. & 21.3{\tiny$\pm$0.6} & 42.0{\tiny$\pm$0.1} & 44.3{\tiny$\pm$0.5} &  45.5{\tiny$\pm$0.1}& \textbf{45.6{\tiny$\pm$0.3}} \\
\bottomrule
\end{tabular}
}

\end{table}

\subsection{Decision Boundary Visualization.}
\begin{figure}[h]

\centering
\includegraphics[width=\linewidth]{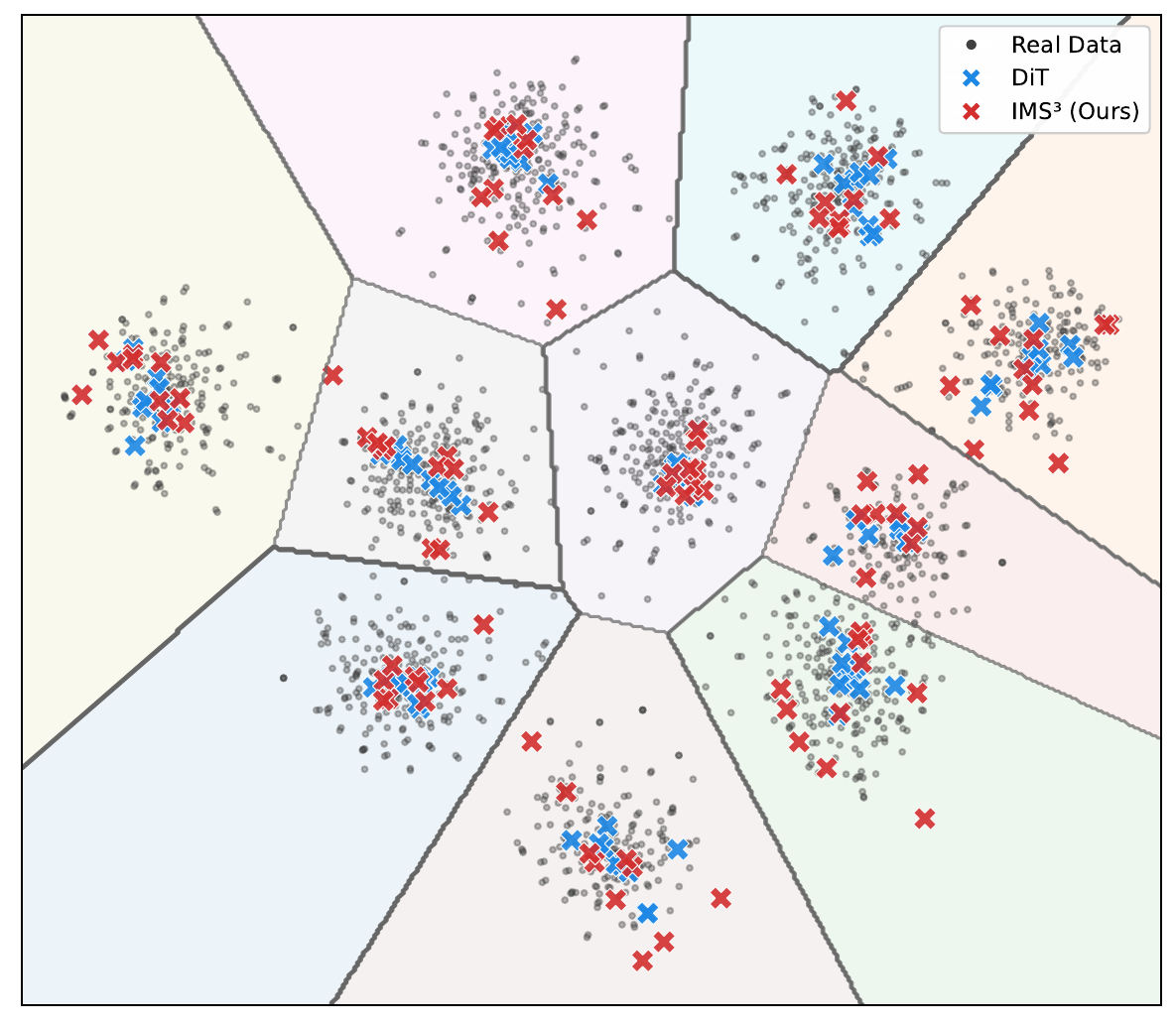}
\caption{Feature-space visualization of distilled samples.}
\label{fig:boundary}

\end{figure}

We visualize classifier features using t-SNE in \Cref{fig:boundary}. Samples selected by IMS$^3$ are closer to the decision boundaries.

\subsection{Real Data Accessibility}
S³ can approximate class centroids using samples generated by the diffusion model itself. As shown in \Cref{tab:gen_cen}, using generated samples to estimate centroids still effective compared to the baselines.

\begin{table}[h]

\centering
\caption{Gen-data centroid results on ImageWoof}
\label{tab:gen_cen}
\resizebox{\linewidth}{!}{
\begin{tabular}{lcccc}
\toprule
 & DiT & Minimax & \textbf{IMS$^3_{\text{Gen}}$}
 & \textbf{IMS$^3_{\text{Real}}$} \\
\midrule
IPC=10 & 34.7{\tiny$\pm$0.5} & 35.7{\tiny$\pm$0.3} & \underline{39.6{\tiny$\pm$0.4}} & \textbf{41.8{\tiny$\pm$0.3}}\\
IPC=50 & 49.3{\tiny$\pm$0.2} & 54.4{\tiny$\pm$0.6} & \underline{55.5{\tiny$\pm$0.3}} & \textbf{57.3{\tiny$\pm$0.5}}\\
\bottomrule
\end{tabular}
}

\end{table}

\section{Visualization}

\begin{figure*}[p]
    \centering
    \hspace*{-2.5cm}
    \includegraphics[scale=2.8]{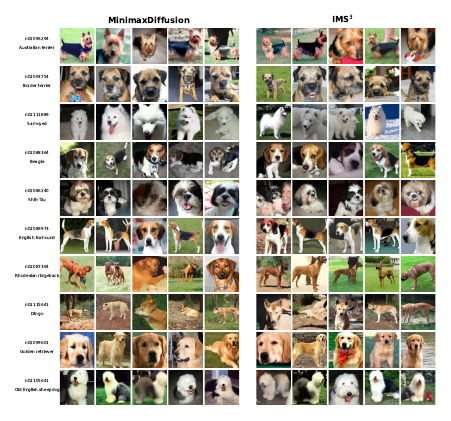}
    \vspace{0.3cm}
    \caption{Comparison between samples selected by Minimax~\cite{minimax} (left) and generated by the proposed \ims  (right) for ImageWoof. The class names are marked at the left of each row.}
    \label{fig:woof111}
\end{figure*}

\begin{figure*}[p]
    \centering
    \hspace*{-2.5cm}
    \includegraphics[scale=2.8]{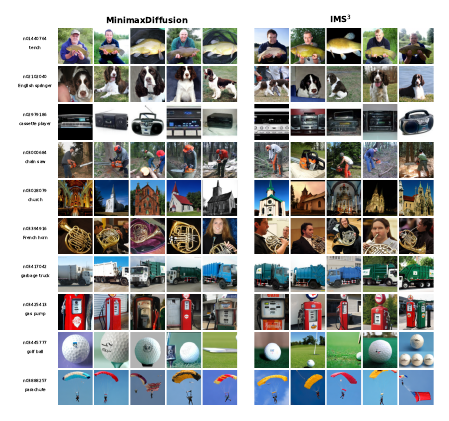}
    \vspace{0.3cm}
    \caption{Comparison between samples selected by Minimax~\cite{minimax} (left) and generated by the proposed \ims  (right) for ImageNette. The class names are marked at the left of each row.}
    \label{fig:nette111}
\end{figure*}

\begin{figure*}[p]
    \centering
    \hspace*{-2.5cm}
    \includegraphics[scale=2.8]{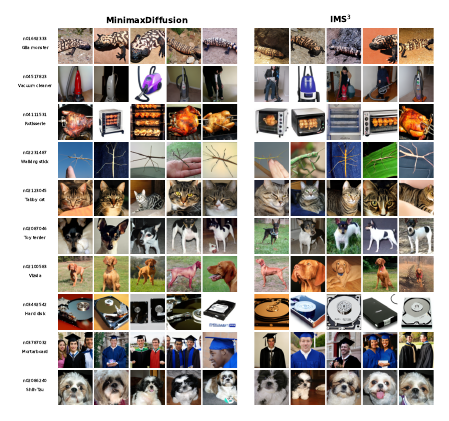}
    \vspace{0.3cm}
    \caption{Comparison between samples selected by Minimax~\cite{minimax} (left) and generated by the proposed \ims  (right) for ImageNet-100 classes 0-9. The class names are marked at the left of each row.}
    \label{fig:100-1}
\end{figure*}

\begin{figure*}[p]
    \centering
    \hspace*{-2.5cm}
    \includegraphics[scale=2.8]{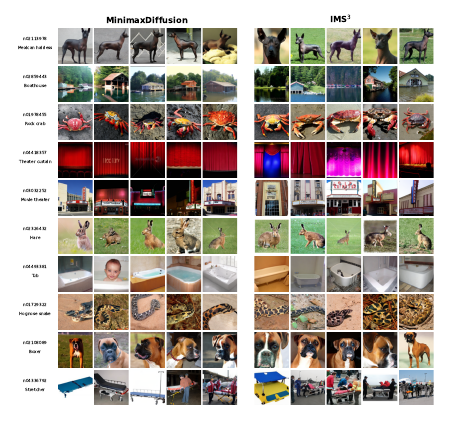}
    \vspace{0.3cm}
    \caption{Comparison between samples selected by Minimax~\cite{minimax} (left) and generated by the proposed \ims  (right) for ImageNet-100 classes 10-19. The class names are marked at the left of each row.}
    \label{fig:100-2}
\end{figure*}

\begin{figure*}[p]
    \centering
    \hspace*{-2.5cm}
    \includegraphics[scale=2.8]{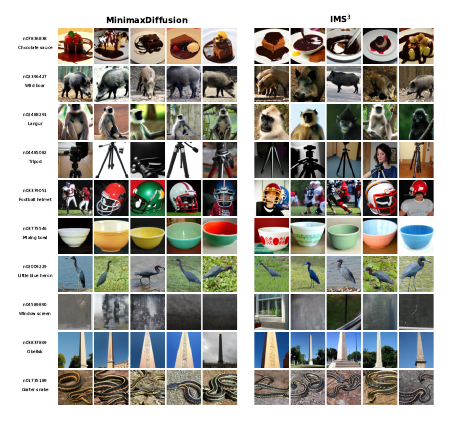}
    \vspace{0.3cm}
    \caption{Comparison between samples selected by Minimax~\cite{minimax} (left) and generated by the proposed \ims  (right) for ImageNet-100 classes 20-29. The class names are marked at the left of each row.}
    \label{fig:100-3}
\end{figure*}

\begin{figure*}[p]
    \centering
    \hspace*{-2.5cm}
    \includegraphics[scale=2.8]{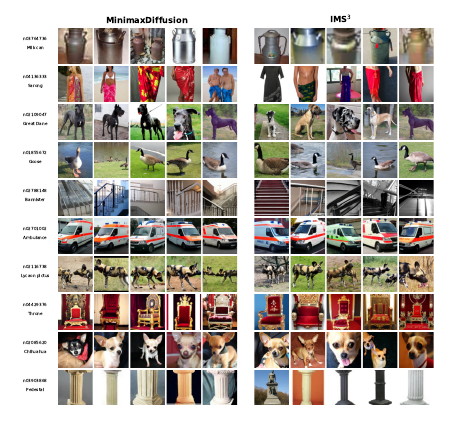}
    \vspace{0.3cm}
    \caption{Comparison between samples selected by Minimax~\cite{minimax} (left) and generated by the proposed \ims  (right) for ImageNet-100 classes 30-39. The class names are marked at the left of each row.}
    \label{fig:100-4}
\end{figure*}

\begin{figure*}[p]
    \centering
    \hspace*{-2.5cm}
    \includegraphics[scale=2.8]{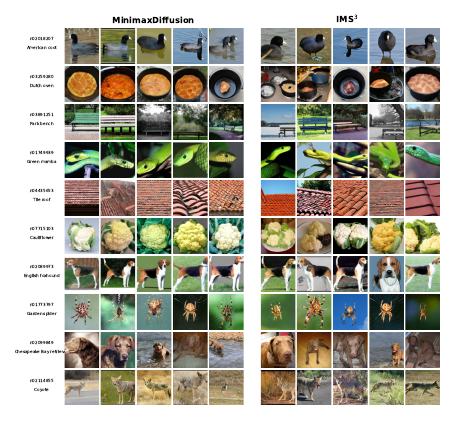}
    \vspace{0.3cm}
    \caption{Comparison between samples selected by Minimax~\cite{minimax} (left) and generated by the proposed \ims  (right) for ImageNet-100 classes 40-49. The class names are marked at the left of each row.}
    \label{fig:100-5}
\end{figure*}

\begin{figure*}[p]
    \centering
    \hspace*{-2.5cm}
    \includegraphics[scale=2.8]{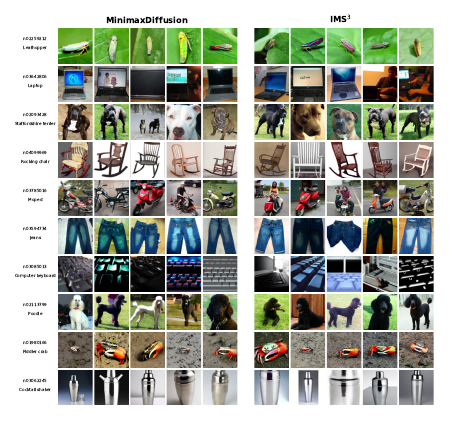}
    \vspace{0.3cm}
    \caption{Comparison between samples selected by Minimax~\cite{minimax} (left) and generated by the proposed \ims  (right) for ImageNet-100 classes 50-59. The class names are marked at the left of each row.}
    \label{fig:100-6}
\end{figure*}

\begin{figure*}[p]
    \centering
    \hspace*{-2.5cm}
    \includegraphics[scale=2.8]{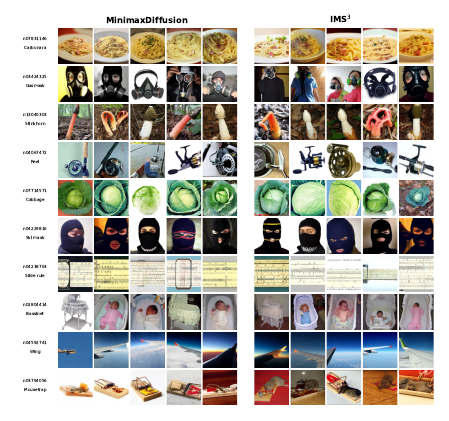}
    \vspace{0.3cm}
    \caption{Comparison between samples selected by Minimax~\cite{minimax} (left) and generated by the proposed \ims  (right) for ImageNet-100 classes 60-69. The class names are marked at the left of each row.}
    \label{fig:100-7}
\end{figure*}

\begin{figure*}[p]
    \centering 
    \hspace*{-2.5cm}
    \includegraphics[scale=2.8]{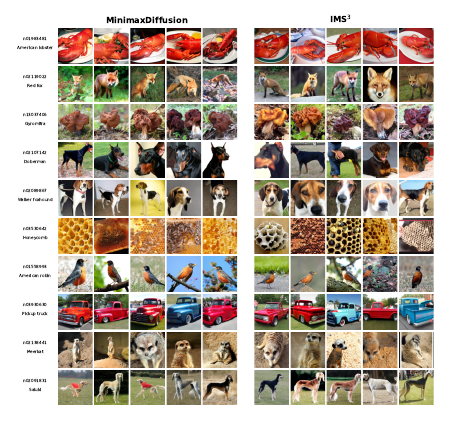}
    \vspace{0.3cm}
    \caption{Comparison between samples selected by Minimax~\cite{minimax} (left) and generated by the proposed \ims  (right) for ImageNet-100 classes 70-79. The class names are marked at the left of each row.}
    \label{fig:100-8}
\end{figure*}

\begin{figure*}[p]
    \centering
    \hspace*{-2.5cm}
    \includegraphics[scale=2.8]{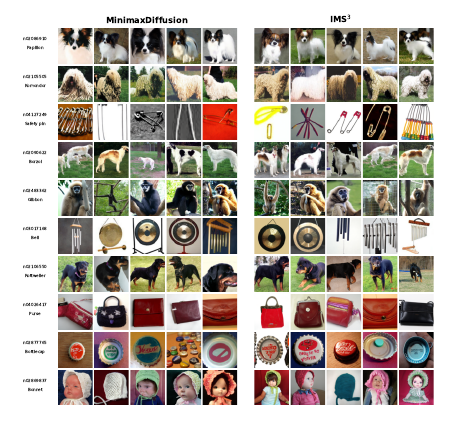}
    \vspace{0.3cm}
    \caption{Comparison between samples selected by Minimax~\cite{minimax} (left) and generated by the proposed \ims  (right) for ImageNet-100 classes 80-89. The class names are marked at the left of each row.}
    \label{fig:100-9}
\end{figure*}

\begin{figure*}[p]
    \centering
    \hspace*{-2.5cm}
    \includegraphics[scale=2.8]{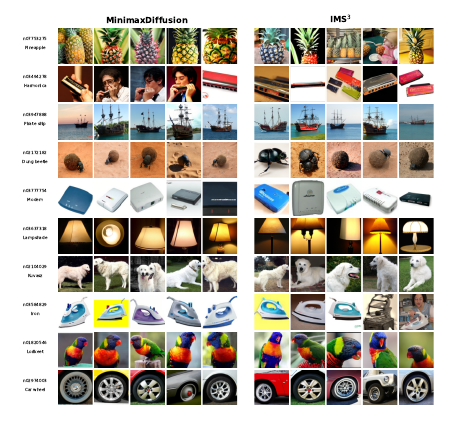}
    \vspace{0.3cm}
    \caption{Comparison between samples selected by Minimax~\cite{minimax} (left) and generated by the proposed \ims  (right) for ImageNet-100 classes 90-99. The class names are marked at the left of each row.}
    \label{fig:100-10}
\end{figure*}

We demonstrate the samples selected by Minimax~\cite{minimax} and \ims\,, \cref{fig:woof111} is the distilled samples from ImageWoof, \cref{fig:nette111} is the distilled samples from ImageNette, and \cref{fig:100-1,fig:100-2,fig:100-3,fig:100-4,fig:100-5,fig:100-6,fig:100-7,fig:100-8,fig:100-9,fig:100-10} are the distilled samples from ImageNet-100.

\end{document}